\documentclass[preprint, 12pt]{elsarticle}

\usepackage{bm}
\usepackage{amssymb}
\usepackage{amsmath}
\usepackage{lineno}
\usepackage{multirow}
\usepackage{graphicx}
\usepackage{xcolor}
\usepackage{algorithm}
\usepackage{algpseudocode}
\usepackage{booktabs}
\usepackage{tabularx} 
\usepackage{CJKutf8} 
\journal{ISPRS Journal of Photogrammetry and Remote Sensing}
\usepackage[colorlinks, citecolor=red]{hyperref}
\begin{document}
  \begin{frontmatter}

    \title{\textcolor{black}{LLM-Driven Training-free Location-Attribute Synergic Fusion: A Closed-Loop Paradigm for Dual-source Encrypted POIs and LULC Mapping}}

    \author[label1]{Chang Li\corref{cor1}} 
    \author[label1]{Xingtao Peng}
    \author[label2]{Yongjun Zhang}
    \author[label1]{Yinfei He} 
    \author[label1]{Cairun Huang}
    \affiliation[label1]{organization={Key Laboratory for Geographical Process Analysis \& Simulation of Hubei Province, and College of Urban and Environmental Sciences},
    addressline={Central China Normal University}, city={Wuhan}, postcode={430000}, state={HuBei}, country={China}}
    \affiliation[label2]{organization={School of Remote Sensing and Information Engineering},
    addressline={Wuhan University}, city={Wuhan}, postcode={430000}, state={HuBei}, country={China}}
    \cortext[cor1]{e-mail: lcshaka@126.com, lichang@ccnu.edu.cn}
    \begin{abstract}
\textcolor{black}{Dual-source encrypted points of interest (DSEP) (i.e., POIs from two encrypted coordinate systems) suffer from two intertwined uncertainty issues, namely location uncertainty (e.g., misalignment caused by nonlinear systematic distortions) and attribute uncertainty (e.g., naming inconsistency), which severely hinder downstream geospatial applications such as large-scale land-use/land-cover (LULC) mapping in encrypted coordinate environments.}
To the best of our knowledge, this paper is the first to propose a large language model (LLM)-driven training-free location-attribute synergic \textcolor{black}{closed-loop joint optimization paradigm based on location-attribute interdependent fusion.
Our paradigm is built upon a closed-loop iterative optimization framework that co-optimizes location fusion and attribute matching.} \textcolor{black}{Attribute-synergic location fusion first employs an LLM-driven attribute matching method (training-free, reducing time complexity from $O(N^2)$ to $O(N)$)} to establish correspondences between DSEP coordinates, then refines the transformation coefficients through an improved PSO algorithm within ISODATA-clustered local subregions. \textcolor{black}{Location-synergic attribute fusion} subsequently reassesses attribute confidence conditioned on the updated geometric residuals via an LLM-fuzzy method, resolving DSEP attribute uncertainty.
\textcolor{black}{The refined attribute correspondences feed back into the next round of location optimization, forming a bidirectional closed-loop that iteratively co-optimizes both the coordinate transformation and the correspondence matrix.}
\textcolor{black}{Sample purification and adaptive contraction of the local search radius drive the iterative loop to essentially converge within two iterations.}
\textcolor{black}{Furthermore, we propose a joint LLM- and encrypted map-driven training-free LULC mapping method that directly inherits land use classification from encrypted maps via location fusion, producing vector-raster integrated LULC mapping training-free results.}
Besides, we further propose \textcolor{black}{a reference-free POI fusion evaluation method} to evaluate our paradigm across 31 provincial capitals and municipalities in mainland China, and the experiments demonstrate that our method outperforms the open-source baseline and SOTA with a DSEP location fusion average residual of \textcolor{black}{4.58} meters and an attribute fusion accuracy of 95.12\%, indicating improvements of \textcolor{black}{1.77} meters and 14.87\%, respectively.
\textcolor{black}{ The LULC mapping achieves an average Macro-F1 of 83.11\% and an average mIoU of 75.14\%, demonstrating effective alignment of encrypted vector data with WGS-84 reference data. Overall}, our method pioneers the location-attribute synergic DSEP fusion with a low-cost (sparse POIs), high-accuracy, highly automated, and training-free solution in a realistic and challenging scenario (i.e., two encrypted maps). \textcolor{black}{The derived correction formula enables georeferencing of encrypted vector data to WGS-84 without field-surveyed GCPs, directly supporting remote sensing product production.}
    \end{abstract}




  \end{frontmatter}


  \section{Introduction}
  \label{sec_1} 
\begin{figure}[h]
    \centering
    \includegraphics[width=0.8\textwidth]{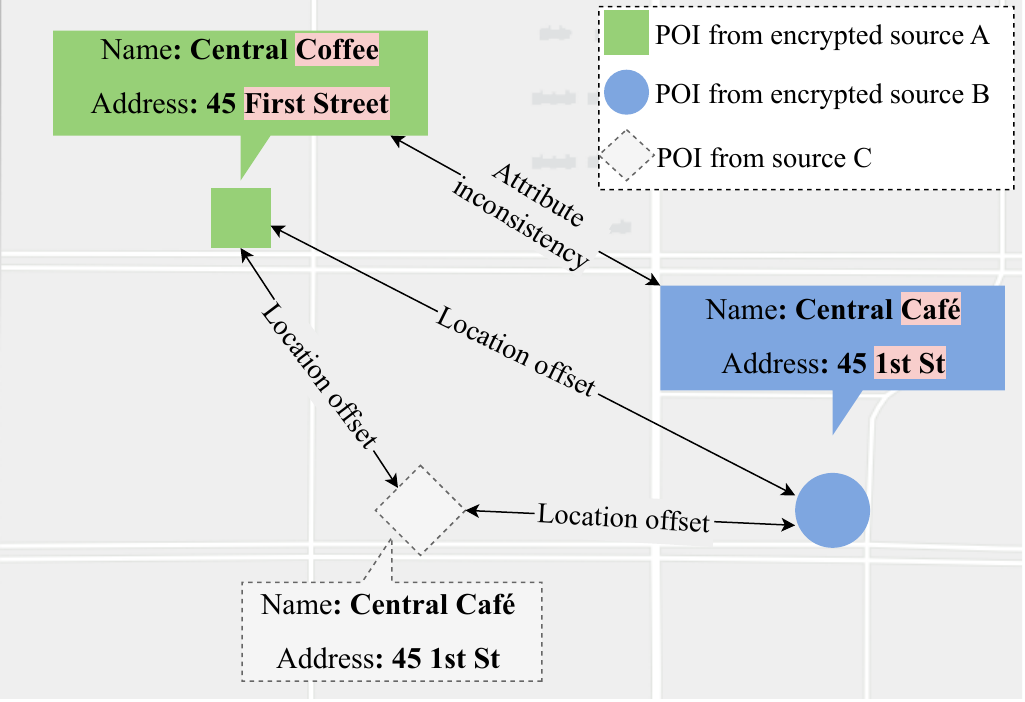}
    \caption{Illustration of coordinate offset and attribute inconsistency in \textcolor{black}{dual-source encrypted points of interest (DSEP)} fusion.
    Although POI attributes or locations are different in multiple sources, location and attribute are interdependent for the same geographic entity and can synergically support each other in the DSEP fusion.}
    \label{intro_1}
  \end{figure}

Points of interest (POIs) represent real-world geographic entities with both location and attribute data \cite{sunConflatingPointInterest2023}.
To enhance data completeness, it is necessary to fuse POI data from multiple sources.
\textcolor{black}{The integration of POI datasets originating from different sources enables the utilization of their heterogeneous attribute information, with each dataset compensating for the attribute biases of others \cite{sunConflatingPointInterest2023}.
POI data have been extensively applied across geographical research domains, including urban spatial structure analysis and human activity modeling \cite{yaoSensingSpatialDistribution2017a, songAreAllCities2018a}.
Although notable advancements have been achieved in the quantitative fusion of multi-source spatial data \cite{yuSelfconsistentDeepGeometric2024}, the qualitative consolidation and accurate alignment of encrypted POIs continue to pose unresolved difficulties.}
\textcolor{black}{A significant challenge is that fusing such multi-source POI data remains difficult, especially when both sources are encrypted coordinate systems.}
We define this task as the dual-source encrypted POI (DSEP) fusion.
As shown in Figure \ref{intro_1}, this challenge manifests as attribute inconsistency and location offset between corresponding POIs (i.e., the same geographic entity).
As two fundamental components of a geographic entity, attributes and location exhibit \textcolor{black}{a location-attribute interdependent fusion relationship}.
This inherent interdependence allows for a synergic fusion, where location can synergize attribute fusion, and conversely, attributes can synergize location fusion.
Despite the growing use of DSEP in practical applications, research on addressing the DSEP location-attribute synergic fusion challenge remains lacking.

A significant challenge in DSEP fusion arises from the proprietary and encrypted nature of the coordinate systems used by major digital map platforms.
For instance, Baidu Maps and Amap are widely used digital maps that provide location-based services in China.
\textcolor{black}{Together, these two platforms serve over one billion users and maintain POI databases exceeding 100 million entries, all under encrypted coordinate systems.
Fusing these two sources is essential for applications ranging from urban planning to autonomous driving, yet the encryption-induced nonlinear offsets (up to hundreds of meters) and attribute ambiguities make existing unencrypted-coordinate methods inapplicable.
This problem affects the entire Chinese geospatial ecosystem and any international application that requires accurate POI data in China.}
Amap employs the GCJ-02 encrypted coordinate system, whereas Baidu Maps uses the BD-09 coordinate system, which is further obfuscated based on GCJ-02.
The geographic data they provide, including POI data, all use encrypted coordinate systems, which primarily manifests in the form of \textcolor{black}{nonlinear systematic distortions} applied to coordinates without any official API for converting to general coordinate systems (e.g., WGS-84). \textcolor{black}{This study focuses on fusing two encrypted coordinate systems (BD-09 and GCJ-02) rather than fusing encrypted with WGS-84 data, for two reasons. First, Chinese platforms such as Amap and Baidu Maps offer superior temporal currency and richer semantic attributes compared with WGS-84 sources available in China. Second, fusing two independently encrypted coordinate systems is inherently more challenging than aligning an encrypted system with WGS-84, because the deviations of both systems must be jointly resolved. For the ground truth validation experiment, the open-source Foursquare Open Source Places POI dataset was used rather than Google POI data, because the former is freely accessible whereas the latter is commercially restricted and closed-source. Because the decryption algorithms of BD-09 and GCJ-02 are inaccessible and no public dataset provides strict one-to-one correspondences between encrypted and WGS-84 POIs, direct accuracy validation on real data is infeasible. Therefore, a synthetic encrypted dataset was adopted for ground truth validation rather than direct comparison of real encrypted and non-encrypted POIs.}
While an open-source method for transforming between these coordinate systems is available \cite{wuDeviationChinaMap}, the reverse-engineered solution often lacks the accuracy required for fusing small-scale geographic entities, though it can serve as a valuable baseline for comparison in an experiment.

The other challenge in DSEP lies in the ambiguous expression of attribute text.
This issue arises when attributes such as names and addresses of the same geographic entity exhibit semantic equivalence yet differ in textual representation across various data sources (Table \ref{tab_intro}).
Differences in data collection methods between Baidu Maps and Amap contribute to these discrepancies.
Typical discrepancies include mismatched number formats, alternative abbreviations, synonym replacement, variations in word order, and abbreviation differences.
\textcolor{black}{Traditional text similarity methods such as edit distance rely on character or word matching, but unstructured and non-standard POI attributes frequently cause mismatches and large errors.}
Leveraging pre-trained geographic knowledge and semantic reasoning, LLMs address these challenges more effectively than traditional methods.
They understand the logical rules of addresses, identify synonymous terms, and assess semantic similarity, which makes them a sound and scalable solution for attribute matching of DSEP.
\textcolor{black}{Deep learning has shown strong performance in POI matching, but its use for training-free POI registration has not yet been explored.}
\begin{table}[htbp]
    \centering
    \caption{Semantic differences in the representation of the same POI between Baidu Maps and Amap.}
    \footnotesize
    \begin{CJK*}{UTF8}{gbsn}
    \begin{tabularx}{\linewidth}{p{4cm}XX}
    \toprule
    \textbf{Attribute of Location} & \textbf{Baidu Maps} & \textbf{Amap} \\
    \midrule
    Name & 北京工业大学附属中学(双桥分校) & 北京工大附中双桥分校 \\
    English Name & Beijing University of Technology Affiliated High School (Shuangqiao Branch) & Affiliated High School of BJUT, Shuangqiao Branch \\
    Address & 双桥中路32号院 & 双桥中路32号院 \\
    English Address & No. 32 Courtyard, Shuangqiao Middle Road & No. 32 Courtyard, Shuangqiao Middle Road \\
    Coordinate (longitude, latitude) & 116.596075, 39.899951 & 116.589449, 39.894216 \\
    \bottomrule
    \end{tabularx}
    \end{CJK*}
    \label{tab_intro}
\end{table}
Therefore, addressing the data fusion challenges caused by encrypted map coordinates in an efficient, automated, and accurate way requires further study.
Based on the review above, \textcolor{black}{three core challenges are identified:}

  \begin{enumerate}[(1)]
\item \textcolor{black}{Joint location-attribute fusion under encrypted coordinates.}
\textcolor{black}{The encryption-induced offsets are nonlinear and spatially heterogeneous, so a single global transformation cannot correct local shifts, while ambiguous attributes cause mismatches and gross errors in correspondence establishment.}
\textcolor{black}{Existing studies address location fusion and attribute fusion separately, and how to exploit their interdependence for joint optimization within a computationally affordable scope remains to be explored.}

\item \textcolor{black}{Automated reference-free evaluation without training data.}
\textcolor{black}{Deep learning based methods require labeled data for training and ground truth for evaluation, which does not apply to automated pipelines or large-scale DSEP fusion, so an evaluation method that requires neither training data nor ground truth is needed to monitor fusion quality at each stage.}

\item \textcolor{black}{Downstream transfer to remote sensing applications.}
\textcolor{black}{Deep learning based land-use/land-cover (LULC) mapping depends on large-scale annotated datasets, whereas encrypted electronic maps contain fine-grained and frequently updated land parcels, and how to exploit them for training-free LULC mapping under encrypted coordinate systems requires further study.}
  \end{enumerate}

To address these challenges, this study proposes a novel closed-loop joint optimization paradigm based on LLM for DSEP fusion, which fuses attribute and location in a synergic approach.
This approach uses a global-to-local framework in which attribute matching and location optimization alternate and feed back into each other, forming a bidirectional closed-loop that iteratively co-optimizes both stages until convergence.
Attributes are first used to roughly match POIs across maps, guiding the initial alignment of their spatial positions.
The refined coordinate transformation then re-conditions attribute confidence, and the updated correspondences feed back into the next round of location optimization, progressively purifying the correspondence set and contracting the local search radius.

\textcolor{black}{In practice, Chinese platforms such as Amap and Baidu Maps update their POI and vector data at high frequency with rich semantic attributes, yet the encrypted coordinate systems introduce systematic positional biases relative to WGS-84 imagery, making coordinate unification a prerequisite for reliable land-use/land-cover (LULC) mapping.
Meanwhile, deep learning based LULC mapping relies on large-scale annotated datasets and extensive model training \cite{baoVisionMambaRemote2025, zhangBridgingSemanticsGeometry2025}, whereas encrypted electronic maps contain fine-grained, frequently updated land parcels with built-in semantic labels, offering an attractive source for training-free LULC mapping.
Therefore, enabling accurate alignment between encrypted vector data and remote sensing imagery is a key prerequisite for exploiting these electronic maps in large-scale automated LULC mapping, and this practical demand motivates the proposed DSEP fusion framework, which unifies heterogeneous coordinate systems and enables efficient vector--raster integration for downstream LULC applications.}

The innovation and contribution of this paper are as follows:
  \begin{enumerate}[(1)]
\item \textcolor{black}{At the paradigm level, this paper proposes a closed-loop joint optimization paradigm for the location-attribute interdependent fusion of DSEP.}
\textcolor{black}{Unlike existing research that processes location fusion and attribute fusion independently in an open-loop pipeline, the paradigm exploits the interdependence between location and attribute, in which attribute matching, location optimization, and attribute reassessment alternate in a bidirectional feedback loop that progressively purifies the correspondence set and drives the joint objective to converge within two iterations.}
\textcolor{black}{Within this paradigm, the LLM-driven attribute matching (LAM) method matches names and addresses through multi-round dialogues within ISODATA-partitioned subregions without any training, reducing the time complexity from $O(N^2)$ to linear level, the global-to-local location fusion (GTLLF) method optimizes the transformation coefficients through an improved particle swarm optimization (PSO) algorithm with adaptive search-space initialization, and the LLM-fuzzy method reassesses attribute confidence conditioned on the updated geometric residuals.}
\textcolor{black}{The paradigm is training-free, reference-free, and highly automated, and it is model-agnostic in that any attribute similarity model can serve as the matching initializer.}

\item \textcolor{black}{At the evaluation level, this paper proposes a reference-free POI fusion evaluation method that requires neither ground truth for accuracy evaluation nor labeled data for training.}
\textcolor{black}{The method reflects relative accuracy at each stage of the fusion process, establishing a highly automated DSEP fusion and metric framework without any manual annotation and supporting large-scale DSEP fusion applications.}

\item \textcolor{black}{At the application level, this paper proposes a joint LLM- and encrypted POI-driven training-free LULC (LEPF-LULC) mapping method that enables vector-raster integrated LULC mapping under encrypted coordinate systems with sparse POIs.}
\textcolor{black}{Unlike segmentation-network-based approaches that depend on large-scale annotated training data and substantial computational resources, this method directly inherits land use classification from encrypted maps via the correction formula, requires no training, and achieves high accuracy with sparse POIs.}
  \end{enumerate}

\textcolor{black}{The remainder of this paper is organized as follows. Section 2 reviews related work, Section 3 presents the closed-loop joint optimization paradigm together with its constituent methods, Section 4 reports the experimental results, Section 5 discusses the optimization mechanism, efficiency, evaluation validity, and limitations, and Section 6 concludes this paper.}

  \section{Related work}
  \label{sec_2}

  \subsection{Large language model}
Currently, ChatGPT and DeepSeek are among the most popular LLMs, representing general and domain-specific models.
ChatGPT is built on the Transformer architecture \cite{brownLanguageModelsAre2020} for natural language tasks.
DeepSeek \cite{deepseek-aiDeepSeekV3TechnicalReport2025} uses a Mixture-of-Experts design and multi-head latent attention, optimized for Chinese.
LLMs have been applied in geosciences \cite{wangGPTLargeLanguage2024, zhangBBGeoGPTFrameworkLearning2024, zhangGeoGPTAssistantUnderstanding2024}.
\textcolor{black}{This work employs LLMs for POI fusion, using ChatGPT and DeepSeek-Chat to distinguish ambiguous POI names and addresses and match identical entries.}
  
  \subsection{Location fusion of POIs}
Discrepancies in coordinates can result from differences in DSEP collection methods and other factors.
Researchers have extensively studied ways to reduce misalignment during POI integration.
\citet{zhaoPoiPointEntity2022} proposed a multi-feature POI fusion approach using spherical distance, edit distance, and the Jaro-Winkler method, which outperformed methods based only on name similarity or spatial proximity.
\citet{cousseauLinkingPlaceRecords2021a} noted that duplicate POIs across sources often differ in location, but nearby POIs are more likely to represent the same entity.
They introduced a geographic encoder that computes Haversine distances and encodes them for further analysis.
Merging, rather than fusing, multi-source POIs is also common, particularly to expand datasets, such as consolidating restaurant POIs from multiple sources.
  
There has been no reported work specifically addressing attribute-synergic location fusion between encrypted coordinate systems (i.e., maps).
Inspired by prior work but differing from the details of location fusion strategies, this study reduces residual errors through local intelligent optimization after coarse coordinate alignment.
This approach mitigates nonlinear spatial offsets in DSEP and enhances the automation and accuracy of POI fusion.

  \subsection{Attribute fusion of POIs}
Attribute fusion involves integrating POI attributes from multiple sources to provide richer and more complete information.
\citet{liDeepLearningMethod2022} proposed a deep learning-based multi-source POI matching method, first using Chinese word segmentation to build a vocabulary and then training a Word2Vec model to obtain word embeddings.
\citet{maMultisourcePointofinterestMatching2025} developed a matching approach based solely on multi-attribute feature similarity, without relying on spatial information, performing semantic attribute fusion to identify corresponding POIs.
Some studies use edit distance to match POI text attributes by counting the operations needed to transform one string into another \cite{piechAutomaticPointsInterest2020, caiResearchMultisourcePOI2022a}.
Advances in machine learning have led to more sophisticated text similarity methods.
\citet{linDeepLearningArchitecture2020a} combined Word2Vec with an Enhanced Sequential Inference Model (ESIM) for toponym matching, achieving 97\% precision.
\citet{cousseauLinkingPlaceRecords2021a} encoded POI text as embeddings and trained neural networks to predict various attributes.
With the rise of LLMs, larger models provide better generalization.
For example, \citet{xingLocalPOIMatching2022} used BERT to convert POI names into vectors for improved feature representation, and \citet{qiuDeepNeuralNetwork2024} integrated BERT with ESIM to handle character substitution, enhancing toponym matching accuracy and robustness.
The Enhanced Semantic Representation Model \cite{liEnhancedSemanticRepresentation2023} fuses multi-source POI attributes via pretraining followed by fine-tuning.
  
Although these approaches are effective, they require model training, depend on large sample sizes and substantial computational resources, and exhibit limited generalization capability.
To further reduce computational cost and enhance the automation of the fusion process, this paper proposes a training-free, global-to-local framework for attribute matching.
Moreover, attribute ambiguity and conflicts are effectively addressed in our approach through the application of fuzzy mathematics.

  \subsection{Location-and-attribute fusion of POIs}
DSEP is inherently uncertain both location and attribute fusion are essential for accurate integration.
\citet{wangEfficientAlgorithmSpatiotextual2020} proposed a parallel approach that searches for similar POIs in spatial and textual domains simultaneously, combining spatial proximity and textual similarity to identify matched pairs.
The contrastive learning framework MoCo-Ga \cite{qiangMomentumContrastiveLearning2024} enables POI retrieval using cross-modal data, applying momentum contrastive instance discrimination for textual attributes and a contrastive module for geographic locations.
\citet{caiResearchMultisourcePOI2022a} introduced a multi-source POI fusion framework using a novel POI data structure and clustering to integrate data for the same POI.
\citet{almeidaAutomaticPOIMatching2018a} employed an isolation forest to match Factual, Facebook, and Foursquare POIs in New York City, achieving approximately 95\% accuracy, and validated it on a dataset from Porto, Portugal. \citet{liDifferentSourcingPoint2020} proposed a multi-attribute constraint model for matching POIs between Baidu Maps and Amap, improving F1-score and recall by 7.1\% and 0.3\%.
Graph-based methods, including both traditional and deep learning approaches, have also been effective.
\citet{novackGraphbasedMatchingPointsofinterest2018a} represented POIs as nodes and matching probabilities as edges to solve fusion problems using graph algorithms, while \citet{yuSelfconsistentDeepGeometric2024} developed a deep multi-source spatial prediction framework that integrates heterogeneous sensor data without ground truth using a learnable fidelity score and a geo-aware graph neural network, showing strong results on both synthetic and real datasets.
These studies highlight the complementary nature of spatial and non-spatial information, which informs our approach.

\textcolor{black}{Table~\ref{tab:related_compare} summarizes the key differences between this work and existing approaches across four axes: attribute matching method, coordinate registration strategy, uncertainty treatment, and interdependence modeling.
Existing methods treat attribute matching and coordinate registration as independent sequential steps in an open-loop pipeline.
\textcolor{black}{This paper exploits the interdependence between location and attribute for joint optimization, where attribute matching results directly constrain location fusion and the aligned location subsequently guides attribute fusion.}
The framework is model-agnostic: any attribute similarity model (LLM, BERT, or traditional string matching) can serve as the initializer, while the interdependent fusion principle provides the accuracy gain.}

\textcolor{black}{
\begin{table}[htbp]
\centering
\caption{Comparison with existing POI fusion approaches across key differential axes.}
\label{tab:related_compare}
\resizebox{\linewidth}{!}{
\begin{tabular}{p{3.2cm}p{2.8cm}p{2.8cm}p{2.8cm}p{2.8cm}}
\toprule
\textbf{Method category} & \textbf{Attribute matching} & \textbf{Coordinate registration} & \textbf{Uncertainty treatment} & \textbf{Interdependence} \\
\midrule
String matching + global transform \cite{zhaoPoiPointEntity2022, caiResearchMultisourcePOI2022a, piechAutomaticPointsInterest2020} & Edit distance / Jaccard & Global OLS / affine & None & Independent, sequential \\
Deep learning matching \cite{liDeepLearningMethod2022, xingLocalPOIMatching2022, qiuDeepNeuralNetwork2024} & BERT / contrastive & Global / local transform & Implicit (during training) & Independent, sequential \\
Graph-based matching \cite{novackGraphbasedMatchingPointsofinterest2018a} & Graph node matching & Not separately handled & None & Independent \\
Multi-source prediction \cite{yuSelfconsistentDeepGeometric2024} & Not handled & GNN prediction & Learnable fidelity & Independent \\
\textbf{This paper} & LLM / any model & RANSAC + PSO local optimization & Explicit, bidirectional & \textbf{Interdependent, synergic} \\
\bottomrule
\end{tabular}}
\end{table}}

\textcolor{black}{Unlike existing open-loop methods that treat attribute matching and coordinate registration as independent sequential steps (e.g., first match by text similarity, then estimate a global transformation), our framework explicitly models the interdependence between the semantic and geometric domains.}

Although a few existing studies attempt to fuse location and attribute information simultaneously, none of them employed alternating synergic fusion.
We propose attribute-synergic location fusion, followed by location-synergic attribute fusion, which incorporates the interdependence of spatial and semantic information to enhance fusion performance.

\subsection{\textcolor{black}{LULC mapping}}
\textcolor{black}{Land use/land cover (LULC) mapping provides an important foundation for characterizing surface spatial patterns, monitoring the ecological environment, and analyzing three-dimensional urban spatial structures\cite{11509353}. Semantic segmentation networks for LULC mapping have evolved from CNN-based architectures to Transformer-based designs and, more recently, to state space models such as Mamba \cite{baoVisionMambaRemote2025}.
These methods share a common dependence on large-scale annotated training datasets, and their inference speed is bounded by GPU throughput, with time complexity proportional to image resolution $O(H \times W)$.
More recently, vision-language models have been introduced into remote sensing segmentation, yet they still require supervised fine-tuning on text-image pairs and produce only raster output.}

\textcolor{black}{The closed-loop joint optimization paradigm is not limited to POI fusion.
Its core principle of jointly optimizing semantic and geometric uncertainty without labeled training data transfers directly to remote sensing interpretation.
In LULC mapping, the attribute uncertainty becomes the semantic-label uncertainty of vector polygons, and the geometric uncertainty becomes the coordinate offset between encrypted vector data and WGS-84 imagery.
The same interdependent fusion framework resolves both, enabling training-free vector-raster integrated LULC mapping.
This cross-domain transfer demonstrates that the paradigm is a general principle rather than a POI-specific heuristic.}

\textcolor{black}{In contrast, the Joint LLM- and encrypted POI-driven training-free LULC mapping proposed in this paper represents a fundamentally different paradigm. At the module level, it replaces the segmentation network with a coordinate correction formula that directly transforms encrypted vector data to WGS-84, inheriting fine-grained LULC classification from electronic maps without any pixel-level inference. Training requires zero annotated samples, in contrast to the thousands of labeled tiles needed by segmentation-based methods. Inference involves only coordinate transformation and vectorization with $O(N)$ time complexity and no GPU dependency, versus the $O(H \times W)$ raster-scanning cost of segmentation networks. The output is a vector-raster integrated product with vector polygon boundaries, avoiding the boundary discretization errors inherent in raster segmentation.}

  \section{Methodology}
  \label{sec_3}

  \subsection{Research route}
  \begin{figure}[htb]
    \centering
    \includegraphics[width=\linewidth]{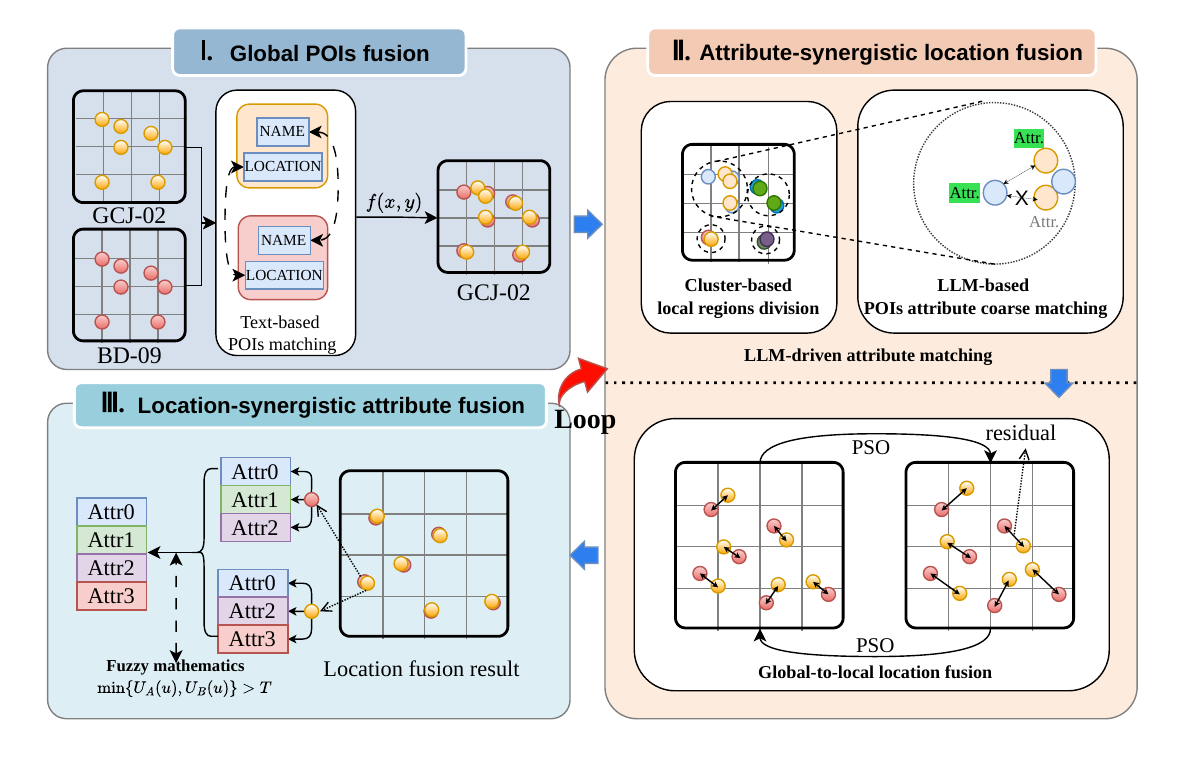}
\caption{Research route.
    The steps of the algorithm are as follows: I.Global POI fusion begins by aligning textual attributes and coordinates to obtain an initial set of fused data.
    II. Local POI fusion is conducted by dividing the data into spatial subregions.
    Within each subregion, POI attributes are preliminarily fused using an LLM.
    III.The intelligent optimization algorithm further reduces location residuals.
    IV. Fuzzy mathematics is employed to resolve attribute ambiguities, resulting in the final fused POI dataset.
    \textcolor{black}{The dashed loop between Attribute-synergic location fusion and Location-synergic attribute fusion indicates the iterative loop: attribute correspondences guide location refinement, and the refined geometry in turn re-conditions attribute confidence, forming a bidirectional feedback that progressively co-optimizes both stages until convergence.}}
    \label{overview}
  \end{figure}

\textcolor{black}{The overall research route of the proposed framework is illustrated in Figure \ref{overview}.
POI data are first acquired through the APIs of Baidu Maps and Amap, and duplicate records with identical names, addresses, and coordinates are removed.
The empirical trigonometric transformation (ETT) based global POI fusion then unifies the two encrypted coordinate systems and provides the initial transformation coefficients and the initial correspondence set.
The core of the framework is a closed-loop that alternates between attribute-synergic location fusion and location-synergic attribute fusion, as indicated by the dashed loop in Figure \ref{overview}.
Within each iteration, the LLM-driven attribute matching establishes correspondences within ISODATA-partitioned local subregions, the global-to-local location fusion optimizes the local transformation coefficients, and the LLM-fuzzy mechanism reassesses attribute confidence conditioned on the updated geometric residuals, generating the final fused POI dataset upon convergence.
The technical details of each module are elaborated in the subsequent subsections, and the closed-loop formulation is presented in Section 3.2.}

  \subsection{\textcolor{black}{Closed-loop Joint Optimization Formulation}}

  \textcolor{black}{This section presents the core methodological contribution of this paper, a closed-loop joint optimization paradigm for DSEP fusion.}
  \textcolor{black}{The DSEP fusion task can be formalized as a joint uncertainty minimization problem.}
  Let $p_i^B$ denote a source POI in the BD-09 coordinate system and $p_j^G$ denote a candidate POI in the GCJ-02 coordinate system.
  \textcolor{black}{The goal is to estimate a coordinate transformation $T_\theta$ parameterized by the trigonometric transformation coefficients $\theta = (a_0, a_1, a_2, a_3, b_0, b_1, b_2, b_3)$ and a one-to-one correspondence matrix $\Pi$ that jointly minimize geometric displacement and attribute mismatch.}

  \textcolor{black}{The attribute uncertainty of a POI pair is defined as the complement of the attribute similarity, given by}
  \begin{equation}
    u_A(p_i, p_j) = 1 - s_A(p_i, p_j), \quad s_A = \min(s_{\mathrm{name}}, s_{\mathrm{address}})
    \label{eq_ua}
  \end{equation}
  \textcolor{black}{where $s_{\mathrm{name}}$ and $s_{\mathrm{address}}$ are the name and address similarity scores respectively.
  The minimum operation follows the fuzzy conjunction principle, requiring both attributes to be consistent for a reliable match.}

  \textcolor{black}{The geometric uncertainty is defined as the normalized residual after coordinate transformation, given by}
  \begin{equation}
    u_G(p_i, p_j) = \frac{\|T_\theta(p_i^B) - p_j^G\|}{r_{\mathrm{local}}}
    \label{eq_ug}
  \end{equation}
  \textcolor{black}{where $r_{\mathrm{local}}$ is the local search radius determined by the three-sigma rule.
  A smaller $u_G$ indicates higher geometric consistency between the transformed source POI and the candidate POI.}

  \textcolor{black}{The joint objective function combines the attribute-weighted geometric residual with an attribute uncertainty penalty, given by}
  \begin{equation}
    \mathcal{L}(\theta, \Pi)^{(k)} = \left[ \sum_{i,j} \Pi_{ij} \cdot s_{A,ij} \cdot \|T_\theta(p_i^B) - p_j^G\|^2 + \lambda \sum_{i,j} \Pi_{ij} \cdot (1 - s_{A,ij})\right]^{(k)}
    \label{eq_joint}
  \end{equation}
  \textcolor{black}{where $\Pi_{ij} \in \{0, 1\}$ enforces one-to-one matching, $\lambda$ balances the geometric and attribute terms, and the superscripts $k$ and $k-1$ denote the current and previous iterations, respectively.
  The first term weights each correspondence by its attribute similarity, granting greater influence to pairs with higher attribute confidence.
  The second term penalizes correspondences with low attribute similarity.}

  \textcolor{black}{
  The loop terminates when
  }
  \begin{equation}
    \left| \mathcal{L}(\theta, \Pi)^{(k)} - \mathcal{L}(\theta, \Pi)^{(k-1)} \right| < \varepsilon
    \label{eq_stop}
  \end{equation}
  \textcolor{black}{
  where $\varepsilon$ is set to $10^{-2}$ in this study, and a maximum iteration count is imposed as a safeguard against non-convergence. This threshold corresponds to centimeter-level precision, which is negligible for map POI location fusion and can be ignored as a residual. Because the geometric term in $\mathcal{L}$ is weighted by the attribute similarity $s_{A,ij} \in [0,1]$, the effective location error entering the objective is lower than the raw geometric residual. Since both the geometric and attribute terms enter $\mathcal{L}$, this criterion reflects the simultaneous convergence of geometry and attribute: each iteration tightens $\theta$ and $\Pi$ and thereby improves the alignment, but the attainable precision is bounded by the residual noise floor, so $\mathcal{L}$ decreases monotonically and converges to that floor rather than improving without limit.
  }

  \begin{algorithm}[!h]
  \color{black}
  \small
  \caption{Closed-loop joint optimization}
  \label{alg_loop}
  \begin{algorithmic}[1]
  \Require BD-09 POIs $\{p_i^B\}$, GCJ-02 POIs $\{p_j^G\}$, tolerance $\varepsilon$, max iteration $K_{\max}$
  \Ensure Transformation $\theta$ and correspondence set $\Pi$
  \State $\theta^{(0)}, \mathcal{P}^{(0)} \gets$ global POI fusion by strict string matching
  \State $r^{(0)} \gets 3\hat{\sigma}^{(0)}$; $\mathcal{L}^{(0)} \gets \sum_{i,j}\Pi_{ij}^{(0)}\left[ s_{A,ij} \| T_{\theta^{(0)}}(p_i^B) - p_j^G \|^2 + \lambda (1 - s_{A,ij}) \right]$
  \For{$k = 1$ to $K_{\max}$}
      \State \textbf{(1) Attribute-synergic location fusion:}
      \State \quad Build candidate pairs within $r^{(k-1)}$; run LLM matching $\rightarrow \Pi^{(k)}_\mathrm{pre}$
      \State \quad PSO update of $\theta^{(k)}$; quality control on non-improving subregions
      \State \textbf{(2) Location-synergic attribute fusion:}
      \State \quad Fuzzy credibility grading; remove low-credibility pairs $\Rightarrow \Pi^{(k)}, \mathcal{P}^{(k)}$
      \State \textbf{(3) Feedback and radius contraction:}
      \State \quad $\hat{\sigma}^{(k)} \gets$ residual dispersion on $\mathcal{P}^{(k)}$; $r^{(k)} \gets 3\hat{\sigma}^{(k)}$
      \State \quad $\mathcal{L}^{(k)} \gets \sum_{i,j}\Pi_{ij}^{(k)}\left[ s_{A,ij} \| T_{\theta^{(k)}}(p_i^B) - p_j^G \|^2 + \lambda (1 - s_{A,ij}) \right]$
      \If{$|\mathcal{L}^{(k)} - \mathcal{L}^{(k-1)}| < \varepsilon$}
          \State \textbf{break}
      \EndIf
  \EndFor
  \State \Return $\theta^{(k)}, \Pi^{(k)}$
  \end{algorithmic}
  \end{algorithm}

  \textcolor{black}{The proposed pipeline implements this formulation through an iterative loop that alternates between Attribute-synergic location fusion and Location-synergic attribute fusion, as depicted by the dashed loop in Figure \ref{overview}.
  Global POI fusion first provides an initial estimate of the transformation coefficients $\theta^{(0)}$ via strict string matching and initializes the correspondence set $\mathcal{P}^{(0)}$ and the local search radius $r^{(0)} = 3\hat{\sigma}^{(0)}$ from the three-sigma rule on global residuals.
  Each subsequent iteration $k$ then executes three ordered steps.}

  \textcolor{black}{
  (1) \emph{Attribute-synergic location fusion} constructs candidate pairs within the current search radius $r^{(k-1)}$, invokes the LLM to identify same-entity correspondences, updates the transformation $\theta^{(k)}$ by PSO on the attribute-weighted geometric residual, and applies a quality-control step (see Section \ref{sec_gtllf_subsec}) that discards subregions whose fitness fails to improve.
  This step tightens $\theta$ because the LLM operates over a smaller candidate pool and PSO optimizes over a cleaner sample support than in the previous iteration.
  }

  \textcolor{black}{
  (2) \emph{Location-synergic attribute fusion} reassesses each remaining pair through the fuzzy-mathematics scheme in Section \ref{sec_35_fuzzy}, which grades the pair by an attribute-similarity credibility level; pairs whose credibility falls below the preset threshold are removed from the correspondence set, yielding the updated $\Pi^{(k)}$ and a contracted sample set $\mathcal{P}^{(k)} \subseteq \mathcal{P}^{(k-1)}$.
  This step tightens $\Pi$ because the previously loose radius admitted platform-inconsistent pairs (same name and address but entity-level positional offset), which the tightened geometry now exposes as low-credibility.
  }

  \textcolor{black}{
  (3) \emph{Feedback and radius contraction.}
  The residual dispersion $\hat{\sigma}^{(k)}$ is recomputed on $\mathcal{P}^{(k)}$, and the search radius contracts to $r^{(k)} = 3\hat{\sigma}^{(k)}$, which sharpens the geometric neighborhood used by step (1) in iteration $k{+}1$.
  This step closes the loop and is the reason accuracy improves iteration by iteration: sample support becomes cleaner, radius becomes tighter, and both $\theta$ and $\Pi$ move onto a smaller and more consistent set of true-correspondence pairs.
  }

  \textcolor{black}{
  The scheme is essentially a block coordinate descent on $\mathcal{L}(\theta, \Pi)$: step (1) decreases $\mathcal{L}$ with $\Pi$ fixed, step (2) decreases $\mathcal{L}$ with $\theta$ fixed, and step (3) monotonically contracts the feasible region.
  Since $\mathcal{L}$ is bounded below and the sequence $\{\mathcal{L}^{(k)}\}$ is monotone non-increasing, the loop converges to a stationary point beyond what any single-pass, open-loop pipeline can achieve.
  The overall procedure is summarized in Algorithm \ref{alg_loop}.
  }

  \subsection{Global POIs fusion}
The global objective function describes the relationship between corresponding POIs.
These POIs are identified automatically through strict string matching, which ensures accuracy but may miss some matches due to attribute ambiguity.
Despite this limitation, the roughly selected pairs make global registration (i.e., fusion) more efficient.
Quadratic polynomials are commonly applied for coordinate rectification, with the X- and Y-directions expressed as follows:
  \begin{equation}
    \begin{cases}
      x=a_{0}+\left(a_{1}X+a_{2}Y\right)+\left(a_{3}X^{2}+a_{4}XY+a_{5}Y^{2}\right) \\
      y=b_{0}+\left(b_{1}X+b_{2}Y\right)+\left(b_{3}X^{2}+b_{4}XY+b_{5}Y^{2}\right)
    \end{cases}
    \label{eq_ols}
  \end{equation}

The formulation of non-rigid registration-based coordinate fusion is expressed as follows:

\begin{equation}
    \begin{array}{l}
E_{\mathrm{tps}}(f) = \sum_{i=1}^{N} \left\| \bm p_i' - f( \bm p_i ) \right\|^2 + \lambda \cdot I_f\\
f(\bm {p}) = \mathbf{A} \bm p + \sum_{i=1}^N \mathbf{w}_i \phi(\|\bm{p} - \bm{p}_i\|) \\
I_f = \iint_{\mathbb{R}^2} \left[ \left( \frac{\partial^2 f}{\partial X^2} \right)^2 + 2\left( \frac{\partial^2 f}{\partial X \partial Y} \right)^2 + \left( \frac{\partial^2 f}{\partial Y^2} \right)^2 \right] dXdY
\end{array}
\label{eq_tpa}
\end{equation}

Where $r_i$ denotes the Euclidean distance between $\bm p_i'=(x,y)$ and $\bm p_i=(X,Y)$, $\bm w_i$ describes the local nonlinear influence of the $i$-th control point on the current coordinate $(X,Y)$.
$\|\cdot\|$ is $L_2$ Norm.
The $\mathbf{A}$ is the coefficient matrix, corresponding to the quadratic polynomial in formula \ref{eq_ols}.
$\phi(r)$ is the TPS kernel function.
To ensure a physically plausible and smooth deformation, the parameters of $f\left(X,Y\right)$ are derived by minimizing the functional of the bending energy function $E_{\mathrm{tps}}$.
  
In addition, the following trigonometric formula \ref{eq_trigonometric} is recommended and is used as one of the comparison methods in subsequent experiments, as it is widely applied in spatial data fusion.
  \footnotesize
{ \begin{equation}\left\{{\begin{array}{*{20}{l}}{x = \;}&{\left\{ {\sqrt{{{(X - {a_0})}^2} + {{(Y - {a_1})}^2}} - {a_2}\sin \left[ {(Y - {a_1}) \cdot \frac{{3000\pi }}{{180}}} \right]} \right\} \cdot }\\ {}&{\cos \left[ {\arctan \left( {\frac{{Y - {a_1}}}{{X - {a_0}}}} \right) - {a_3}\cos (X - {a_0}) \cdot \frac{{3000\pi }}{{180}}} \right]}\\ {y = \;}&{\left\{ {\sqrt{{{(X - {b_0})}^2} + {{(Y - {b_1})}^2}} - {a_2}\sin \left[ {(Y - {b_1}) \cdot \frac{{3000\pi }}{{180}}} \right]} \right\} \cdot }\\ {}&{\sin \left[ {\arctan \left( {\frac{{Y - {b_1}}}{{X - {b_0}}}} \right) - {b_3}\cos (X - {b_0}) \cdot \frac{{3000\pi }}{{180}}} \right]}\end{array}}\right.
\label{eq_trigonometric}\end{equation} }
  
Here, $(x,y)$ denotes the reference coordinates (longitude and latitude of POIs from Amap), while $(X,Y)$ represents the transformed coordinates (longitude and latitude of POIs from Baidu Maps).   \textcolor{black}{ $3000\pi/180$ is a unit conversion coefficient that transforms longitude/latitude degrees to radians scaled by the Mercator projection factor of the encrypted system.}
The coefficients $a_{i}, {\ b}_{i}(i=0,\ 1,\ 2,\ 3,\ 4,\ 5\ldots)$ are the correlation parameters of the global objective function, with the optimal values determined by stepwise regression.
This formula is applied not only for global registration but also for local registration.
In addition, the random sample consensus (RANSAC) \cite{fischlerRandomSampleConsensus1981} algorithm is employed to remove mismatches and retain only reliable POI pairs between Baidu and Amap.
Global POIs fusion (GPF) quickly aligns the coordinates of DSEP with its speed advantage, giving a foundation that still contains minor residuals and therefore requires further refinement in the subsequent location fusion process.
\textcolor{black}{In the closed-loop joint optimization paradigm, GPF provides the initial transformation $\theta^{(0)}$ and the initial correspondence set $\mathcal{P}^{(0)}$, which serve as the inputs to Algorithm \ref{alg_loop} in Section 3.2.}

  \subsection{Attribute-synergic location fusion}
  \textcolor{black}{This subsection implements step (1) of the iterative loop in Section 3.2 and is repeated in each iteration with a progressively contracted search radius, rather than executed only once.}
  \subsubsection{Cluster-based local regions division}
Since POIs are unevenly distributed, this randomness affects registration accuracy.
To reduce residual error, the global area is divided into local spatial subregions for DSEP fusion.
This partitioning enables more accurate location fusion within each subregion.
ISODATA \cite{ballISODATANovelMethod1965a}, an adaptive clustering algorithm, achieves this by iteratively splitting clusters with high variance and merging those with low variance, thus automatically determining suitable subregions.
This adaptive mechanism has relatively low parameter sensitivity and enhances robustness.
  \textcolor{black}{Parameters were set following classic recommendations for urban spatial clustering, calibrated to produce subregions of approximately 10--15 km, matching the spatial scale of local encryption distortions.}

  \subsubsection{Local search circle}
To address attribute ambiguity in fusion, this paper applies a local search circle strategy, inspired by Tobler's First Law of Geography that "near things are more related than distant things" \cite{toblerComputerMovieSimulating1970a}.

  \begin{figure}[h]
    \centering
    \includegraphics[width=0.6\linewidth]{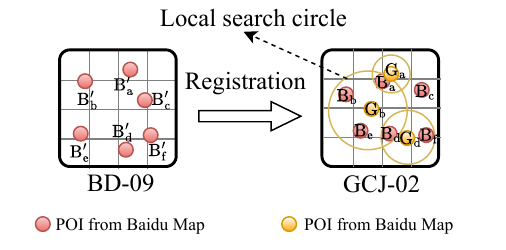}
    \caption{The local POI matching algorithm based on the local search circle}
    \label{local_search}
  \end{figure}

After GPF, BD-09 POIs are transformed into GCJ-02.
Due to encrypted coordinates, identical entities still show inconsistent locations after transformation, as shown in Figure \ref{local_search}.
The residual error is therefore used to estimate the local search circle.
In this paper, the standard deviation serves as the measure of uncertainty.
If residuals follow a normal distribution, values outside $[-3\sigma,3\sigma]$ are treated as outliers.
The initial search circle defines potential POI matches, since POI pairs are formed from data within this circle.
To further improve matching accuracy, we set the maximum radius $R$ as the criterion, calculated as:

{ \footnotesize \begin{equation}R=\sum_{i=1}^{n}{\left\| p_{i}^\text{B}-p_{i}^\text{D} \right\|}/n+3\cdot \sqrt{\frac{1}{n-1}\sum\limits_{i=1}^{n}{{{\left( \left\| p_{i}^\text{B}-p_{i}^\text{D} \right\|-\sum\limits_{i=1}^{n}{\left\| p_{i}^\text{B}-p_{i}^\text{D} \right\|}/n \right)}^{2}}}}\label{eq_buffer}\end{equation} }

where $n$ is the number of the corresponding POIs pair in each search circle; $\textbf{p}_{i}^{\text{B}}$ and $\textbf{p}_{i}^{\text{D}}$ are the coordinate pairs of Baidu Maps POI and Amap POI, respectively.

After GPF, in each local search circle, Baidu POIs $\mathrm{B_a^\prime}$, $\mathrm{B_b^\prime}$, $\mathrm{B_c^\prime}$, $\mathrm{B_d^\prime}$ and $\mathrm{B_e^\prime}$ become $\mathrm{B_a}$, $\mathrm{B_b}$, $\mathrm{B_c}$, $\mathrm{B_d}$ and $\mathrm{B_e}$.
Among them, $\mathrm{B_a}$, $\mathrm{B_b}$ and $\mathrm{B_d}$ correspond to Amap POIs $\mathrm{G_a}$, $\mathrm{G_b}$, and $\mathrm{G_d}$.
Their names and addresses are similar but not identical. For example, if $\mathrm{B_a}$, $\mathrm{B_b}$, $\mathrm{B_c}$, $\mathrm{B_d}$ and $\mathrm{B_e}$ all fall into the local search circle of $G_b$, then four candidate pairs $\left(\mathrm{G_b},\mathrm{B_a}\right)$, $\left(\mathrm{G_b},\mathrm{B_b}\right)$, $\left(\mathrm{G_b},\mathrm{B_d}\right)$, and $\left(\mathrm{G_b},\mathrm{B_e}\right)$ are generated.
These candidate pairs are then matched using LLMs, which effectively narrows the search scope and resolves attribute ambiguity.
To this end, we design attribute fusion methods based on two different LLMs: ChatGPT-based and DeepSeek-based.
By lowering the number of candidate POIs per query, the local search circle reduces overall complexity from $O(N^{2})$ to $O(K\cdot N)$, leading to lower LLM token usage and improved matching accuracy.

  \subsubsection{LLM-driven attribute matching}
\begin{figure}[t]
    \centering
    \includegraphics[width=0.25\linewidth]{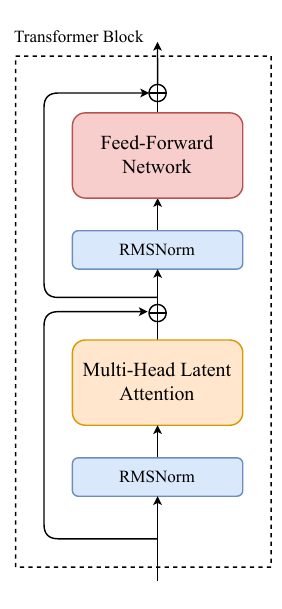}
    \caption{The basic architecture of DeepSeek-Chat\cite{deepseek-aiDeepSeekV3TechnicalReport2025}, which is similar to the transformer block and consists of multiple blocks composed of multi-head latent attention (MLA), a DeepSeekMoE-based feedforward network, and root mean square layer normalization (RMSNorm).}
    \label{DS_overview}
  \end{figure}

As shown in Table \ref{tab_intro}, the attributes and location in DSEP are both opposition and unity.
Specifically, the attributes and locations of an entity are two distinct characteristics (i.e., opposition).
Simultaneously, once the attributes are determined under multiple attribute constraints, the location of the entity is also determined (i.e., unity), and vice versa.
Based on this principle, this paper leverages different pre-trained LLMs to match the semantic attributes of DSEP at low cost, eliminating the need for training.
Specifically, this paper directly integrates the official open APIs of both DeepSeek (https://api-docs.deepseek.com/) and ChatGPT (https://platform.openai.com/docs/overview) into our pipeline, enabling automated and batch processing for POI attribute matching, which gives the advantage of zero training cost compared to traditional supervision-based methods such as BERT.
ChatGPT is a closed-source model, whereas DeepSeek-Chat (Figure \ref{DS_overview}) is an open-source LLM featuring a 671B-parameter sparse MoE architecture.
It enables dynamic resource allocation, which improves computational efficiency and reduces costs.

Unlike DeepSeek-R1, DeepSeek-Chat is a general-purpose LLM primarily designed for text understanding and dialogue, whereas R1 is tailored for dense inference applications.
A key innovation in DeepSeek is the low-rank joint compression of key-value (KV) pairs in the MLA\cite{deepseek-aiDeepSeekV3TechnicalReport2025}.
Specifically, the down-projection matrix $\mathbf{W}^{DKV}$ is used to compress the embedding vector $\bm{h}_{t}$ of the $t$-th token into a low-dimensional latent vector $\bm{c}_{t}^{KV}$:
  \begin{equation}
    \bm{c}_{t}^{KV}={\mathbf{W}^{DKV}}{\bm{h}_t}
  \end{equation}
Then, the compressed latent vectors are reconstructed into the value matrix $\bm {v}_{t}^{C}$ and key matrix $\bm{k}_{t}^{C}$ using the up-projection matrices $\mathbf{W} ^{UK}$ and $\mathbf{W}^{UV}$, respectively.
Finally, $\mathbf{W}^{KR}$ is the matrix used to produce the decoupled key $\bm{k}_{t}^{R}$ that carries rotary positional embedding (RoPE)\cite{sunMassiveActivationsLarge2024}:
  \begin{equation}
    \begin{array}{l}
      \bm{k}_t^C = {\mathbf{W}^{UK}}\bm{c}_t^{KV}                                \\
      \bm{v}_t^C = {\mathbf{W}^{UV}}\bm{c}_t^{KV}                                \\
      \bm{k}_t^R = \mathrm{RoPE}\left( {{\mathbf{W}^{KR}}{\mathbf{h}_t}} \right)
    \end{array}
  \end{equation}
The low-rank joint compression allows MLA to maintain performance comparable to standard attention while significantly reducing cache usage and cost.

This paper leverages LLM-driven attribute matching (LAM), which uses two mainstream LLMs, DeepSeek-Chat and ChatGPT, to adaptively and efficiently achieve consistent semantic attribute matching in DSEP.
Both ChatGPT and DeepSeek perform POI attribute matching via APIs combined with predefined prompts.
Specifically, the LLM is guided by predefined prompts that establish the conversational context and model role, while dialogue templates with input and output rules ensure consistent responses.
First, the semantic environment and decision rules are incorporated into the LLM.
Then, conversation templates containing both questions and answers are designed to improve the model's understanding of the POI matching task.
Table \ref{table1} provides the predefined prompts and sample dialogues used as inputs to the LLM.
This paper conducts batched multi-round dialogues with the LLM and collects the formatted matching results.

  \begin{table}[!h]
    \centering
\caption{Predefined prompts.
    By incorporating environment prompts, judgment rules, and predefined multi-round dialogues, the LLM produces structured outputs, enabling efficient aggregation of multi-source POI attribute matches.}
    \resizebox{\linewidth}{!}{%
    \begin{tabular}{p{4cm}p{15cm}}
      \hline
      Type of prompts & Content of prompts\\
      \hline
      Environment Prompts & In subsequent interactions, I will provide two formatted Chinese addresses, separated by an underscore. You are required to determine purely based on semantics whether the two addresses refer to the same location and respond with a \textcolor{black}{confidence score between 0 and 1 (where 0 indicates they are definitely not the same address and 1 indicates they are definitely the same address).} \\
      Judgment Rules & Determine whether the addresses refer to the same location based on the following principles: If a larger-scale location contains a smaller-scale location, they should be considered different locations. For instance, "Beijing International Studies University" and "Beijing International Studies University-Qinxue Building" should be identified as different locations. If the descriptions differ but the addresses are semantically identical, they should be recognized as the same address. For example, "No. 1, Lane 3, Dong'an Street" and "No. one, Third Lane, Dong'an Street, Fengtai Town" should be considered the same address. Please note that I will provide multiple groups of addresses (different groups will be separated by "\textbar{}"). Please respond in the same sequence and format accordingly. \\
      \multirow{2}{4cm}{\small Manually set the content of the dialog 1} & \textbf{Q}: "Beijing International Studies University"\_"Beijing International Studies University"\textbar{}"Building 307, Yard 28, Guangqu Road"\_"Building 307, Yard 28, Guangqu Road" \\& \textbf{A}: \textcolor{black}{0.98\textbar{}0.95 }\\
      \multirow{2}{4cm}{\small Manually set the content of the dialog 2} & \textbf{Q}: "Beijing International Studies University"\_"Beijing Foreign Studies University"\textbar{}"No. 2, Wenxing Street"\_"Near No. 2, Wenxing Street" \\ & \textbf{A}: \textcolor{black}{0.02\textbar{}0.93} \\
      \hline
    \end{tabular}
    }
    \label{table1}
  \end{table}

\subsection{Global-to-local location fusion}
\label{sec_gtllf_subsec}
\textcolor{black}{The PSO optimization in GTLLF realizes the parameter update $\theta^{(k)}$ in step (1) of the iterative loop, minimizing the attribute-weighted geometric residual that corresponds to the first term of the joint objective in formula \ref{eq_joint}.}
This paper leverages a global-to-local location fusion (GTLLF), which guides PSO \cite{kennedyParticleSwarmOptimization1995a} to progressively optimize the initial parameters of the global transformation function.
The parameter estimation problem is reformulated as a gradient-free optimization task, which avoids the complexity of nonlinear map registration models.
To optimize the accuracy of local DSEP fusion, we apply the PSO algorithm as follows:
  \begin{equation}
    \bm{v}_{ij}\left(t+1\right)=\omega\bm{v}_{ij}\left(t\right)+c_{1}r_{1}\left[\bm
    {P}_{\text{best}}\left(t\right)-\bm{a}_{ij}\left(t\right)\right]+c_{2}r_{2}\left[\bm{G}_{\text{best}}\left(t\right)-\bm{a}_{ij}\left(t\right)\right]
  \end{equation}
  
  \begin{equation}
    \bm{a}_{ij}\left(t+1\right)=\bm{a}_{ij}\left(t\right)+\bm{v}_{ij}\left(t+1\right)
  \end{equation}
Here, $t$ denotes the number of iterations, $i$ is the particle index, and $j$ is the dimension.
The inertia weight $\omega$ is non-negative: a larger $\omega$ strengthens the search ability for the optimal coefficient.
$c_1$ and $c_2$ are positive acceleration coefficients, where $c_1$ adjusts individual particle behaviour and $c_2$ controls social behavior.
$r_1$ and $r_2$ are random numbers uniformly distributed in $[0,1]$.
The coefficients $(a_0,\ a_1,\ a_2,a_3,\ldots)$ and $(b_0,\ b_1,b_2,\	b_3\ ,\ldots)$ in the $x$- and $y$-directions represent the particle's basic form in PSO.
For example, $\bm{a}_{ij}=\left(a_{i0},\ a_{i1},\ a_{i2},a_{i3},\ldots\ldots\right)$ represents the position of particle i, while $\bm{v}_{ij}=\left(v_{i0},\ v_{i1},\	v_{i2},\ v_{i3},\ldots \right)$ denotes its velocity.
PSO proceeds iteratively.
Each particle retains a memory of its best position so far,$\bm{P}_\text{best}=\left(P_{i0},\ \ P_{i1},\ \ P_{i2},\ \ P_{i3}\ ,\ldots\ldots\ \ \right)$, as well as the global best position found across the entire swarm, $\bm{G}_\text{best}=(G_0,\ G_1,\ G_{2\ },\ G_3\ ,\ldots)$.

For local POI regression using formula \ref{eq_ols}, the objective functions of particle $i$ in the $x$- and $y$-directions are defined as formula \ref{eq8}: {\footnotesize \begin{equation}\begin{cases}\min\left(f_{ix}\right)=\min\left[\sum_{j=1}^{n}\left|a_{i0}+\left(a_{i1}X_{j}+a_{i2}Y_{j}\right)+\left(a_{i3}{X_j}^{2}+a_{i4}{X_j}Y_{j}+a_{i5}{Y_j}^{2}\right)-x_{j}\right|\right]\\ \min\left(f_{iy}\right)=\min\left[\sum_{j=1}^{n}\left|b_{i0}+\left(b_{i1}X_{j}+b_{i2}Y_{j}\right)+\left(b_{i3}{X_j}^{2}+b_{i4}{X_j}Y_{j}+b_{i5}{Y_j}^{2}\right)-y_{j}\right|\right]\end{cases} \label{eq8}\end{equation} }\textcolor{black}{These objective functions operationalize the geometric term of the joint objective in formula \ref{eq_joint}, where each residual contribution is weighted by the attribute similarity $s_{A,j}$ as shown in formula \ref{eq_weighted}.}
  
For thin-plate spline (TPS) (formula \ref{eq_tpa}), the primary goal of the optimization is to find the best correction vector $\Delta \mathbf{A}$ that minimizes the total energy functional $E_\mathrm{tps}$, $\mathbf{A}_0$ is the initial parameter set: {\footnotesize
  \begin{equation}
\mathop {\min }\limits_{\Delta \mathbf{A} } E_\mathrm{tps} (\Delta \mathbf{A} ) = \sum\limits_{i = 1}^n \left[| |\bm p_i^\prime  - f(\bm{p_i},{\mathbf{A} _0} + \Delta \mathbf{A} )|{|^2} + \lambda  \cdot {I_f}({\mathbf{A} _0} + \Delta \mathbf{A} ) \right]\\
 \label{eq8-1}\end{equation} }
  
   For the global transformation function using the formula \ref{eq_trigonometric}, the objective functions are:

    \begin{equation}
\label{eq9}
\begin{aligned}
\min(f_{ix})&= \min\sum_{j=1}^{n}\Bigg| \bigg\{ \sqrt{(X_j - a_0)^2 + (Y_j - a_1)^2}- a_{2}\sin\left[ (Y_{j}- a_{1}) C_{\mathrm{trig}}\right]\bigg\} \cdot\\&\quad \cos\bigg[\arctan \left( \frac{Y_j - a_1}{X_j - a_0}\right)- a_{3}\cos\left(X_{j}- a_{0}\right) C_{\mathrm{trig}}\bigg] - x_{j}\Bigg| \\ 
\min(f_{iy})&= \min \sum_{j=1}^{n}\Bigg| \bigg\{ \sqrt{(X_j - b_0)^2 + (Y_j - b_1)^2}- b_{2}\sin\left[ (Y_{j}- b_{1}) C_{\mathrm{trig}}\right] \Bigg\}\cdot\\&\quad \sin\bigg[ \arctan\left(\frac{Y_j - b_1}{X_j - b_0}\right) - b_{3}\cos(X_{j}- b_{0}) C_{\mathrm{trig}}\bigg] - y_{j}\Bigg|
\end{aligned}
\end{equation}
  \textcolor{black}{The term $C_{\mathrm{trig}} = 3000\pi/180$ is a unit conversion coefficient that transforms longitude/latitude degrees to radians scaled by the Mercator projection factor of the encrypted system.}

  \textcolor{black}{The attribute similarity $s_A$ from the preceding LLM matching stage enters the PSO objective function as a fixed weighting factor, implementing the first term of the joint objective in formula \ref{eq_joint}.
  It is important to note that PSO optimizes the trigonometric transformation coefficients $\theta = (a_0, a_1, a_2, a_3, b_0, b_1, b_2, b_3)$ that define the coordinate transformation $T_\theta$, not the parameters of the LLM.
  The LLM produces $s_A$ as a pre-computed similarity score that remains fixed during the PSO optimization process.
  In the LLM-based matching path, $s_A$ takes a continuous confidence value in $[0,1]$ returned by the LLM, so correspondences with higher confidence exert greater influence on the geometric optimization of $\theta$.
  This confidence weighting follows the general weighted formulation in formula \ref{eq_joint}, where $s_A$ can also take continuous values from soft-coded similarity scores produced by BERT.
  The attribute-weighted objective function is expressed as}
  \begin{equation}
    \min(f_{ix}^{\mathrm{w}}) = \min\sum_{j=1}^{n} s_{A,j} \cdot \| T_\theta^{(x)}(X_j, Y_j) - x_j \|
    \label{eq_weighted}
  \end{equation}
  \textcolor{black}{where $s_{A,j}$ is the attribute similarity of the $j$-th POI pair and $T_\theta^{(x)}$ denotes the $x$-component of the trigonometric transformation in formula \ref{eq_trigonometric}.
  Pairs with higher attribute confidence exert greater influence on the optimization, anchoring the transformation to the most reliable correspondences.
  This weighting scheme embeds attribute information directly into the geometric optimization, realizing the synergic principle described in the joint objective.}

  This study uses the coefficients of the global transformation function as the initial parameters for PSO, providing high-accuracy starting values. \textcolor{black}{The key improvement over standard PSO lies in the search-space initialization strategy: instead of manually assigning a uniform search range to all parameters, the search range for each transformation coefficient is automatically determined by three standard deviations ($3\sigma$) around the corresponding global coefficient, yielding parameter-specific bounds that reflect the actual distribution of each coefficient.} The optimization is then performed within this narrower parameter range determined by the three-sigma rule 
  $a_{i0}=\left[a_{0}{-}3\sigma_{a_0},\ a_{0}{+}3\sigma_{a_0}\right]$\allowbreak,
  $a_{i1}=\left[a_{1}{-}3\sigma_{a_1},\ a_{1}{+}3\sigma_{a_1}\right]$\allowbreak,
  $a_{i2}=\left[a_{2}{-}3\sigma_{a_2},\ a_{2}{+}3\sigma_{a_2}\right]$\allowbreak,
  $a_{i3}=\left[a_{3}{-}3\sigma_{a_3},\ a_{3}{+}3\sigma_{a_3}\right]$, $\ldots$\ (i=1,\ 2,\ 3,\ \ldots). This approach searches for the optimal transformation coefficients that minimize the residual defined in formula \ref{eq9}, enabling faster convergence and more accurate solutions. The refinement improves alignment accuracy in locally distorted subregions while reducing parameter sensitivity and the complexity of optimization. If PSO does not sufficiently enhance local accuracy, the search range is gradually reduced until it surpasses the global accuracy or the maximum number of iterations is reached. \textcolor{black}{The specific PSO parameter settings are detailed in the experiment section.}

  \subsection{Location-synergic attribute fusion}
  \label{sec_35_fuzzy}
  \textcolor{black}{This subsection implements step (2) of the iterative loop in Section 3.2: the geometric residuals from GTLLF are used to re-assess attribute confidence and update the correspondence matrix $\Pi^{(k)}$, corresponding to the second term of the joint objective in formula \ref{eq_joint}.}
  After the GTLLF process, \textcolor{black}{nonlinear systematic distortions} are reduced, allowing DSEP to be fused based on attributes. Using LLM-based homonymous matching methods, such as ChatGPT or DeepSeek, similarity results are confidence scores in $[0,1]$, which quantify the degree of matching certainty in the output. To improve the flexibility of the proposed closed-loop joint optimization paradigm, traditional natural language processing (NLP) methods can be applied. \textcolor{black}{NLP-based approaches such as BERT produce soft-coded similarity scores in the form of probabilities. To handle these soft-coded outputs, a fuzzy mathematics fusion mechanism is specifically designed to integrate probabilistic similarity scores from language models, as detailed in the following.}

  \textcolor{black}{As with any training-free classification method, certain initial conditions (e.g., the threshold $T$, ISODATA clustering parameters, and PSO hyperparameters) are preset to default values before execution. Once these defaults are configured, the entire process runs without further human input, and no manual inspection or adjustment of intermediate results is required.}

  \textcolor{black}{The fuzzy mathematics fusion described below can be interpreted as a posterior confidence update conditioned on geometric residuals, thereby implementing the determination of the correspondence matrix $\Pi$ in formula \ref{eq_joint}.
  After GTLLF yields the optimized transformation $T_\theta$, the geometric residual $u_G$ in formula \ref{eq_ug} provides a condition under which the attribute confidence of each POI pair is reassessed.
  Pairs with small geometric residuals retain or strengthen their attribute confidence, since geometric consistency corroborates the attribute match.
  Pairs with large geometric residuals receive attenuated attribute confidence, since geometric inconsistency casts doubt on the attribute match even when name or address similarity appears sufficient.
  This conditional reassessment corresponds to the second term of the joint objective, where the penalty on low attribute similarity is effectively modulated by geometric evidence.
  The fuzzy intersection operation and the thresholded maximum selection that follow operationalize this principle by filtering candidate pairs through both attribute similarity and an implicit geometric consistency check.}

  This paper leverages a SoftMax-of-LLM-based fuzzy mathematics method to calculate the similarity of DSEP name and address. Let $U$ denote the universe of POI pairs and $F$ represent the set of all fuzzy sets on $U$. For the name and address similarity fuzzy sets $s_{\mathrm{name}}, s_{\mathrm{address}} \in F$, the fuzzy intersection is:

  \begin{equation}
    s_{A}\left(u\right)=s_{\mathrm{name}}\left(u\right)\wedge s_{\mathrm{address}}
    \left(u\right)=\min{\left\{s_{\mathrm{name}}\left(u\right),s_{\mathrm{address}}\left(u\right)\right\},\ }
    \ \forall u\in U
  \end{equation}
  where $s_{\mathrm{name}}\left(u\right)$ and $s_{\mathrm{address}}\left(u\right)$ are the similarity scores of name
  and address of POI respectively, which are calculated by SoftMax and take continuous values in $[0,1]$.
  \textcolor{black}{The combined similarity $s_A = \min(s_{\mathrm{name}}, s_{\mathrm{address}})$ defined in formula \ref{eq_ua} is exactly the attribute similarity $s_{A,ij}$ in the joint objective of formula \ref{eq_joint}. It is emphasized that $s_A$ takes continuous values in $[0,1]$, rather than a hard-coded 0 or 1. Furthermore, $s_{A,ij}$ can never vanish to zero: as formulated in \ref{eq7}, only candidate pairs with $s_{A,k}(u) > T$ are retained, where the threshold $T$ is the lower bound given by the uncertainty levels in Table \ref{table2}; hence the selected similarity satisfies $s_{A,ij} = \max\{s_{A,k}(u) \mid s_{A,k}(u) > T\} \in (T, 1]$. Low-credibility pairs below $T$ are discarded by the fuzzy credibility grading in Section \ref{sec_35_fuzzy}, so the retained pairs always carry strictly positive attribute similarity and the attribute-weighted term in formula \ref{eq_joint} never degenerates to zero.}
  In the same search circle, there is a one-to-many situation, and LLM has a prediction error when data has ambiguity. So, the final attribute similarity of the selected POI pair in the same search circle, denoted $s_{A,ij}$, which is exactly the value entering the joint objective in formula \ref{eq_joint}, is calculated as follows:
  \begin{equation}
  s_{A,ij}(u) = \max\Big\{\, s_{A,k}(u) \;\Big|\; s_{A,k}(u) > T,\; k = 1, \ldots, n \,\Big\}
  \label{eq7}
  \end{equation}
  where $s_{A,k}(u)=\min\{s_{\mathrm{name},k}(u), s_{\mathrm{address},k}(u)\}$ is the attribute similarity of the $k$-th candidate pair, $T$ is a certain threshold, and $n$ is the number of POIs in the same search circle.
  \textcolor{black}{The threshold $T$ is determined based on the uncertainty levels defined in Table \ref{table2}, serving as a lower bound for filtering rather than a sensitive tuning parameter.}
  When $s_{A,k}(u)$
  higher than $T$, a POIs pair is considered possibly the corresponding POIs.
  When $s_{A,ij}$ is the maximum value in the same search circle, the POIs pair is regarded
  as the corresponding POIs. In other words, $s_{A,k}(u)
  >T$ is the similarity for a POIs pair, and $s_{A,ij}$ is the final similarity for
  POIs pairs in the same search circle.

  Based on Table \ref{table2}, during the DSEP fusion processing, the input
  dataset consists of $P_{1}$, $P_{2}$, $P_{3}$, where $P_{1}$ and $P_{2}$ are single
  dataset that POIs of Baidu Maps and Amap that doesn't participate in attribute
  matching, $P_{3}$ is POIs pairs that participate in attribute matching. The result
  of the fusion $D$ is obtained according to the following process:
  \begin{enumerate}[(1)]
\item Referring to Table \ref{table2}, POIs pairs with levels (I) and levels (II) are saved as $A$. POIs pairs with levels (III) and levels (IV) are saved as $B$. POIs pairs with levels (V) and levels (VI) are saved as $C$. The LLM confidence scores are normalized into probabilities by the SoftMax function, and ambiguity levels are determined using the thresholds in Table \ref{table2}.

\item A normalized fusion dataset $C_1$ is created from the names and addresses of POIs in $A$. The dataset $C_2$ is obtained by combining $C$ and $C_1$.

\item The initial fusion dataset called $C_{3}$ is obtained by fusing the $P_{1}$ and $P_{2}$.

\item The final fusion dataset called $D$ is obtained by fusing the $C_{2}$ and $C_{3}$
  \end{enumerate}

  \begin{table}[h]
    \centering
\caption{Uncertainty levels of POIs attribute fusion.
    Levels I: Exactly the same, II: Very similar, III: Basically similar, IV: Uncertain, V: Dissimilar, and VI: Completely inconsistent.}
    \resizebox{0.6\linewidth}{!}{
    \begin{tabular}{cccc}
      \hline
      Similarity            & Levels                                   & Similarity         & Levels                                   \\
      \hline
      $1 > s_{A,ij} \ge 0.9$ & \uppercase\expandafter{\romannumeral1} & $0.9 > s_{A,ij} \ge 0.8$ & \uppercase\expandafter{\romannumeral2} \\
      $0.8 > s_{A,ij} \ge 0.6$  & \uppercase\expandafter{\romannumeral3} & $0.6 > s_{A,ij} \ge 0.4$ & \uppercase\expandafter{\romannumeral4} \\
      $0.4 > s_{A,ij} \ge 0.1$  & \uppercase\expandafter{\romannumeral5} & $0.1 > s_{A,ij} \ge 0$  & \uppercase\expandafter{\romannumeral6} \\
      \hline
    \end{tabular}
    }
    \label{table2}
  \end{table}

  {\color{black}
  \subsection{\textcolor{black}{Joint LLM- and encrypted POI-driven training-free LULC mapping}}
  To address the problem that vector data in encrypted coordinate systems cannot be directly overlaid with WGS-84 remote sensing imagery, we propose \textcolor{black}{a Joint LLM- and encrypted POI-driven training-free LULC mapping} method comprising three key steps.

  (1) Coordinate mapping: The corrected WGS-84 coordinates $(x_j, y_j)$ of each vector element are obtained by applying the GTLLF trigonometric transformation (formula \ref{eq9}) with PSO-optimized parameters $\boldsymbol{\theta}=\{a_0, a_1, a_2, a_3, b_0, b_1, b_2, b_3\}$:
  \begin{equation}
    \begin{aligned}
      x_j = \Big\{&\sqrt{(X_j - a_0)^2 + (Y_j - a_1)^2} - a_2\sin\!\left[(Y_j - a_1)\, C_{\mathrm{trig}}\right]\Big\} \cdot\\
      &\cos\!\left[\arctan\!\left(\tfrac{Y_j - a_1}{X_j - a_0}\right) - a_3\cos(X_j - a_0)\, C_{\mathrm{trig}}\right] \\[4pt]
      y_j = \Big\{&\sqrt{(X_j - b_0)^2 + (Y_j - b_1)^2} - b_2\sin\!\left[(Y_j - b_1)\, C_{\mathrm{trig}}\right]\Big\} \cdot\\
      &\sin\!\left[\arctan\!\left(\tfrac{Y_j - b_1}{X_j - b_0}\right) - b_3\cos(X_j - b_0)\, C_{\mathrm{trig}}\right]
    \end{aligned}
    \label{eq_lulc_coord}
  \end{equation}
  where $(X_j, Y_j) \in \mathcal{V}_{\text{GCJ-02}}$ are the GCJ-02 coordinates of the $j$-th vector element and $\boldsymbol{\theta}$ is obtained from GTLLF (formula \ref{eq9}) via PSO optimization. This step achieves transformation from the encrypted coordinate system to WGS-84 without requiring field-surveyed ground control points but relying only on sparse POIs and electronic map vectors, eliminating coordinate offsets.

  (2) Image-vector overlay: The corrected vector polygons with their land use labels are rasterized onto the WGS-84 imagery grid, producing the vector-raster integrated product $\mathcal{M}$:
  \begin{equation}
    \mathcal{M}(u, v) = \begin{cases} l_j, & \text{if } (u, v) \in \mathrm{Poly}\!\left(\{(x_j, y_j)\}\right) \\ \mathcal{I}_{\text{WGS-84}}(u, v), & \text{otherwise} \end{cases}
    \label{eq_lulc_overlay}
  \end{equation}
  where $(u, v)$ is a pixel coordinate in the WGS-84 imagery $\mathcal{I}_{\text{WGS-84}}$, $\mathrm{Poly}(\cdot)$ denotes the polygon region formed by the corrected vector vertices, and $l_j$ is the land use category of the $j$-th vector element. This step precisely overlays the corrected vector data with remote sensing imagery, producing vector-raster integrated mapping results rather than pure raster output.

  (3) \textcolor{black}{Semantic-layer-driven vectorization}: Electronic maps encode land use categories using \textcolor{black}{consistent semantic layers, where polygons sharing the same layer represent the same land use type.}
  The vectorization recovers the semantic label $l_j$ from the color triplet $(\mathrm{RGB})_j = (r_j, g_j, b_j)$ extracted from each vector element of the encrypted electronic map:
  \begin{equation}
    l_j = c^{-1}\!\left[(\mathrm{RGB})_j\right], \quad c^{-1}: \mathbb{R}^{3} \rightarrow \mathcal{C}
    \label{eq_lulc_color}
  \end{equation}
  where $c^{-1}$ maps each RGB color triplet back to its land use category in the category set $\mathcal{C}$, and $c: \mathcal{C} \rightarrow \mathbb{R}^{3}$ is the forward \textcolor{black}{semantic-layer} mapping defined by the encrypted electronic map specification. This step directly inherits the LULC classification information from electronic maps, avoiding boundary errors inherent in segmentation networks. The entire process has a time complexity of $O(N)$, where $N$ is the number of POIs.
  \textcolor{black}{A notable advantage of this approach is that it does not require dense POI distribution.
  Because electronic maps employ a consistent semantic-layer encoding where all polygons of the same layer share an identical land use category, a single POI within any given semantic region suffices to determine the land use classification of all polygons sharing that layer across the entire map.
  The correction formula derived from sparse DSEP fusion thus propagates from individual POI locations to entire vector layers through this semantic-layer correspondence, enabling LULC mapping even in regions where POIs are sparse.}
  }

  \section{Experiment}
  \subsection{Experimental setup}
\begin{figure}[h]
    \centering
    \includegraphics[width=0.8\linewidth]{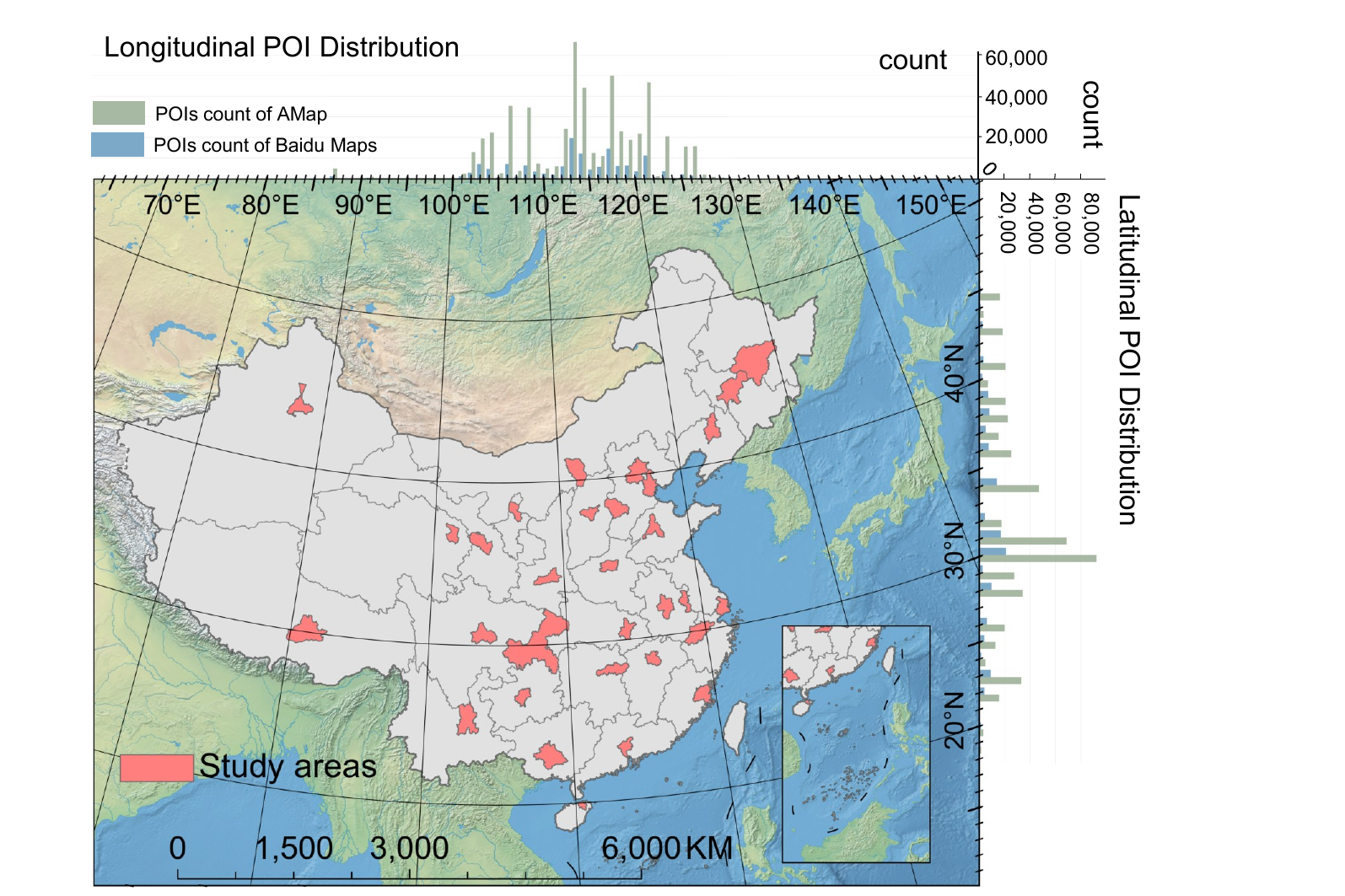}
\caption{Study areas.
    The basemap is based on approval No. GS (2024) 0650 reprojected into an equidistant cylindrical projection to ensure consistent latitude and longitude intervals.}
    \label{study_area}
  \end{figure}

Because neither Baidu Maps nor Amap DSEP data are available for the Taiwan, Hong Kong, and Macao regions of China, these regions were excluded from the study.
As shown in Figure \ref{study_area}, the study area still covers 31 provincial capitals or municipalities across mainland China.
  A total of 135,518 POIs from Baidu Maps and 530,947 POIs from Amap were utilized.
  \textcolor{black}{POI data were collected via official map APIs in 2025, ensuring a consistent temporal baseline across all cities.}

  \subsection{Baseline and evaluation metrics}
  In the comparative experiments, because no prior studies have reported methods for the location-attribute synergic fusion of DSEP, there is no existing baseline that matches the scope of our problem.
  Therefore, for (1) DSEP location fusion, we adopted the only publicly available open-source method as the baseline, and included thin-plate spline and OLS for supplementary comparison.
  For (2) DSEP attribute fusion, we used ChatGPT, DeepSeek, and a fine-tuned BERT model as baselines, since no dedicated encrypted-POI attribute fusion methods have been published.
  Other non-open-source approaches were not compared because they cannot be reproduced or applied to the DSEP setting.

  This paper leveraged \textcolor{black}{a reference-free DSEP fusion evaluation method} for evaluation.
  The core metric was the residual errors between fused coordinates from two encrypted coordinate systems, which was a more realistic and challenging scenario.
  Specifically, this paper computed the residual between transformed coordinates after location fusion (e.g., from Baidu Maps to Amap) as the evaluation metric.
  Therefore, labels or ground truth were not needed to evaluate the fusion results.
  This relative positional discrepancy provided a consistent and practical measure of alignment quality without requiring any absolute ground truth.

  \subsection{The global residual error}
\begin{figure}[htb]
    \centering
    \includegraphics[width=\linewidth]{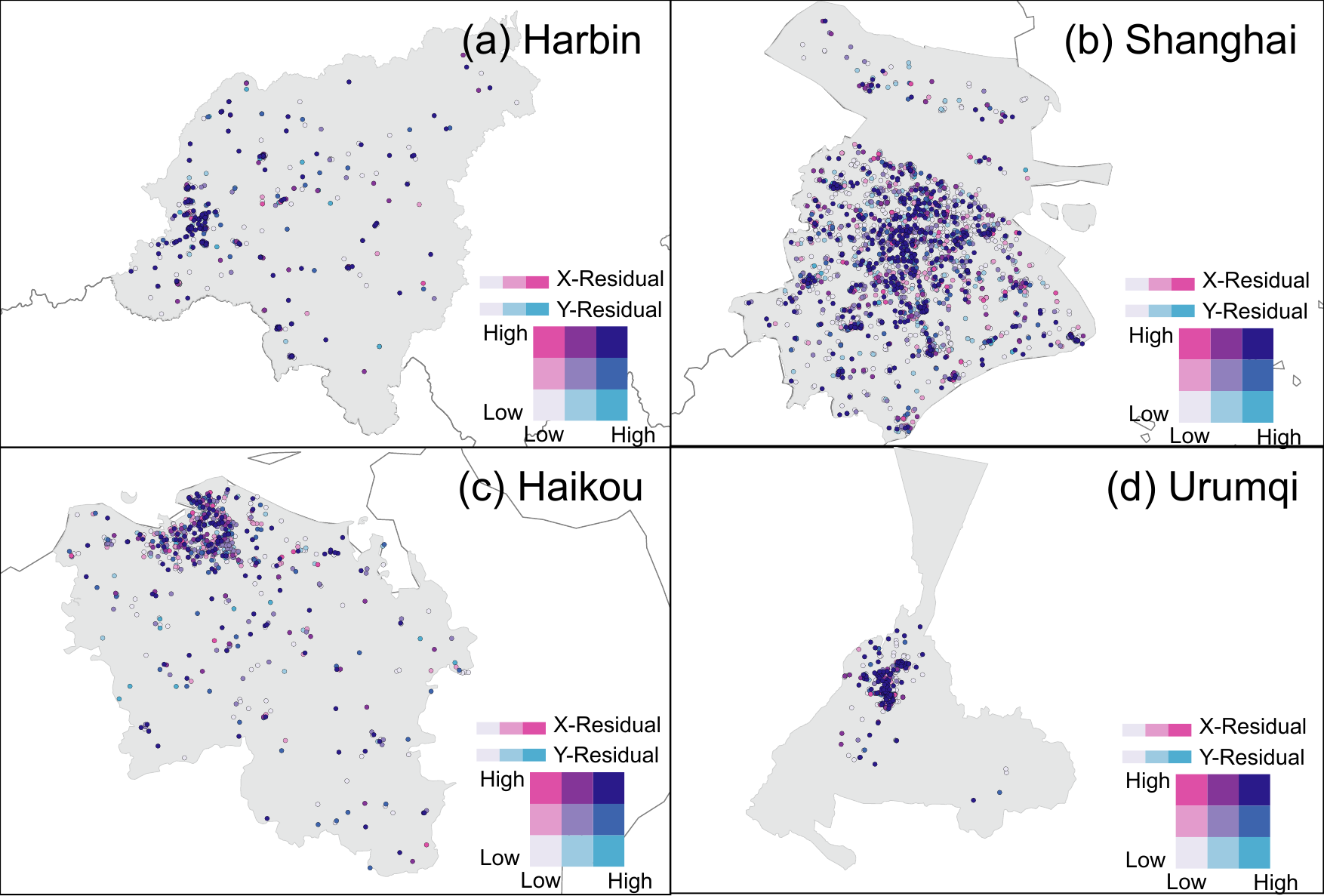}
\caption{Illustration of residual coordinate (geometric) errors in X/Y (longitude and latitude) direction.
    Due to space limitations, subfigures (a), (b), (c), and (d) show residuals for representative cities located in the north, east, south, and west of China (Harbin, Shanghai, Haikou, and Urumqi), respectively, where the randomness of residuals can be found in different directions.}
    \label{G_error}
  \end{figure}

  Through the open-source API, DSEP still has the residual coordinate error. As shown in Figure \ref{G_error}, the residual location error is random in direction and distance.
  In particular, existing open-source location fusion methods cannot perfectly invert the encrypted coordinate transformation, indicating the need for further fine-grained adjustment. 

  \subsection{The results of location fusion}
  Since no prior study has specifically addressed location fusion between DSEP, there is no existing baseline for direct comparison.
  Therefore, this paper designed four comparative methods to verify the effectiveness of the proposed approach: (1) open-source method (empirical trigonometric transformation, ETT); (2)global ordinary least squares (GOLS); (3)first stepwise regression then PSO (FSR-PSO); (4) thin plate spline (TPS).
  \textcolor{black}{The residuals reported in this section were computed over all matched POI pairs rather than a sparse set of ground control points, thereby reflecting the overall spatial consistency of the transformation across each city.}
  \textcolor{black}{Among these methods, TPS produced higher residuals compared with GTLLF.
  Because TPS relies on a global smoothness constraint to minimize bending energy, noise and outliers in the LLM-matched POI pairs propagated through the entire interpolation surface.
  As shown in Table \ref{tab:compare_1}, TPS yielded a mean residual of 15.60 meters for ChatGPT and 15.64 meters for DeepSeek-Chat, indicating that a subset of points was corrected to distant locations.
  This amplification effect arose because the global TPS basis function propagated local errors across the entire transformation field, making it less competitive for encrypted coordinate systems with complex non-linear distortions.}

  Within the LAM framework, DSEP was fused using ChatGPT and DeepSeek as LLMs, respectively. GTLLF was then applied to the matching results of both LLMs.
  In GOLS, coordinates were optimized using ordinary least squares (OLS).
  In FSR-PSO, coordinates were first optimized with the objective function defined in formula \ref{eq8}, and OLS was subsequently applied.
  In TPS, a non-rigid transformation was applied by constructing the canonical TPS system matrix, minimizing the bending energy of the solution while maintaining global affine consistency.
  In GTLLF, optimization relied solely on the objective function in formula \ref{eq9}.
  Stepwise regression and an ill-conditioned matrix adjustment were introduced to address the problems of over-parameterization and instability.
  Since quadratic polynomials may cause overfitting in local regions with limited POIs, hypothesis testing guided the selection of the trigonometric transformation as the global objective function.
  In addition, GTLLF incorporated quality control.
  If the local accuracy after PSO did not exceed that of the ETT after three iterations ($K$=3), PSO was terminated and the raw result was adopted. 
  \textcolor{black}{The PSO parameters were set as follows: cognitive coefficient $c_{1}=0.1$, social coefficient $c_{2}=0.1$, inertia weight $\omega=0.3$, $100$ particles, and $200$ to $300$ iterations. The search space for each parameter was constrained by the three-sigma rule around the global transformation coefficients, ensuring that local refinement stayed within a physically meaningful range. The relatively low cognitive and social coefficients were chosen because the objective of PSO in this framework was not to discover an entirely new transformation from scratch but rather to fine-tune an already well-fitted global equation, where excessive exploration could destabilize the solution.}

  To determine the most suitable objective function for PSO, we conducted comparative experiments using formulas \ref{eq8},\ref{eq8-1} and \ref{eq9} in registration tasks. The results in Table \ref{tab:compare_1} show that formula \ref{eq9} achieved the best performance, effectively capturing nonlinear errors. It provided a sufficiently accurate fit for encrypted coordinate systems (BD-09 and GCJ-02), where the introduction of linear OLS reduced model accuracy. By contrast, the quadratic polynomial (formula \ref{eq8}) proved unsuitable for registering BD-09 and GCJ-02 POI data. As noted earlier, encrypted coordinate systems typically adopt complex, non-formulaic encryption, making a simple quadratic polynomial inadequate for transformation approximation. Even with OLS as a secondary optimization, error reduction was limited, leaving the method less competitive.

\begin{table}[h]
  \centering
  \color{black}
\caption{Location registration error of particle swarm optimization (PSO) based on different objective functions.
    $\mathrm{PSO_{Poly}}$ represents the objective function based on formula \ref{eq8}, while FSR-PSO refers to the process of applying $\mathrm{PSO_{Poly}}$ to the coordinates followed by OLS.
    The same applies to $\mathrm{PSO_{Trig}\text{-}OLS}$ based on the objective function formula \ref{eq9}.}
  \resizebox{0.75\columnwidth}{!}{%
    \begin{tabular}{cccc}
      \toprule
      \multirow{2}{*}{\textbf{LLMs}} & \multirow{2}{*}{\textbf{Algorithm}} & \multicolumn{2}{c}{\textbf{Performance Metrics (m)}} \\
      \cmidrule(lr){3-4}
      & & Mean & Median \\
      \midrule
      & Origin & 1286.74 & 1286.95 \\
      \midrule
      \multirow{6}{*}{ChatGPT}
      & FSR-PSO                             & 888.67             & 941.54  \\
      & $\mathrm{PSO_{Poly}}$                & 598.05             & 577.31  \\
      & $\mathrm{PSO_{Trig}\text{-}OLS}$    & 904.68             & 964.64  \\
      & GOLS                                & 915.28             & 978.8  \\
      & TPS                                 & 15.60  & 8.73    \\
      & \textbf{GTLLF (Ours)}                      & \textbf{4.42}       & \textbf{2.21} \\
      \midrule
      \multirow{6}{*}{DeepSeek-Chat}
      & FSR-PSO                             & 884.45             & 927.45  \\
      & $\mathrm{PSO_{Poly}}$                & 634.24            & 620.04  \\
      & $\mathrm{PSO_{Trig}\text{-}OLS}$    & 900.55             & 956.5  \\
      & GOLS                                & 913.87             & 975.99  \\
      & TPS                                 & 15.64  & 8.68    \\
      & \textbf{GTLLF (Ours)}                      & \textbf{4.48}       & \textbf{2.44} \\
      \bottomrule
    \end{tabular}
  }
  \label{tab:compare_1}
\end{table}

\begin{figure}[htb]
    \centering
    \includegraphics[width=\linewidth]{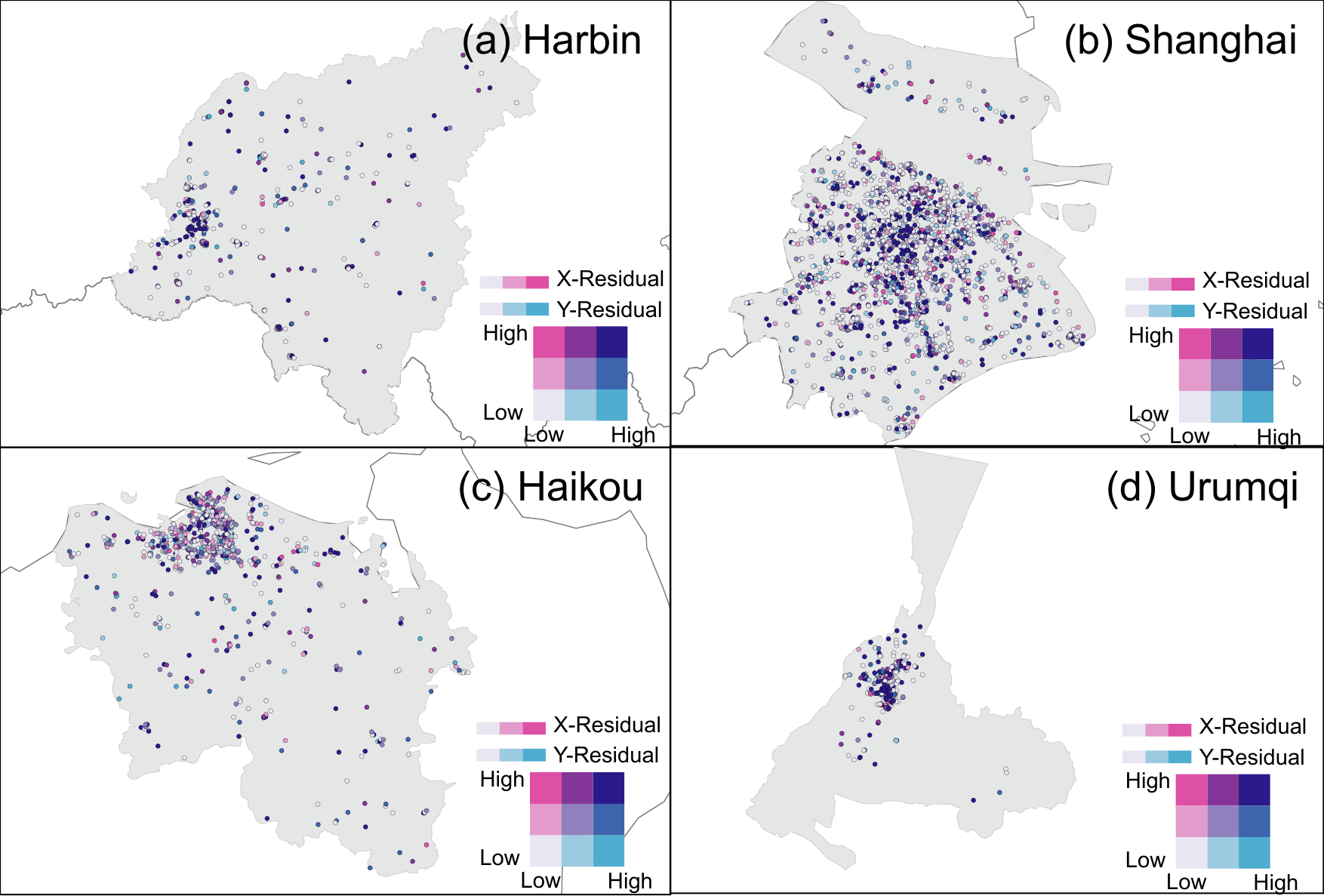}
\caption{Illustration of residual coordinate (geometric) errors in X/Y (longitude and latitude) direction after \textcolor{black}{global-to-local location fusion (GTLLF)}.}
    \label{L_error}
  \end{figure}

  Figure \ref{L_error} shows a significant boost in coordinate stability compared to the ETT findings shown in Figure \ref{G_error}.
  Whereas the ETT method produced scattered and high-magnitude directional shifts, the GTLLF approach yielded a constrained and uniform offset field.
This visual comparison corroborates the quantitative data, confirming that GTLLF effectively mitigates the irregular non-linear distortions that standard ETT methods do not address, ensuring superior positional accuracy across varying geographic regions.
  \textcolor{black}{The progressive global-to-local strategy enhances convergence and reduces spatial residuals, as reflected by the constrained offset field after local PSO refinement.}

  As shown in Table \ref{table_compare_2}, the origin error reveals the magnitude of the non-linear offset without location fusion, with a national average mean cumulative error of approximately 1286.74 meters. Because of the encryption algorithms, combining POI data from the GCJ-02 and BD-09 coordinate systems introduced not only semantic matching errors caused by ambiguous POIs but also coordinate offset errors.
  Using the open-source coordinate conversion method, ETT reduced the offset to some extent.
  GTLLF further decreased the error by an average of 1.77 meters.
  \textcolor{black}{In surveying and mapping, positional accuracy is a long-standing objective metric. Although the gain of 1.77~m is relatively small, it is still appreciable in this field, and meter-level precision matters for many downstream geospatial applications.}
  This improvement can be explained by two factors: GTLLF can correctly identify the same POI among multiple ambiguous candidates, thereby lowering semantic matching errors, and it employs the PSO algorithm to refine the coordinate alignment.
  \textcolor{black}{The faster convergence and reduced spatial residuals achieved by GTLLF, as described in the Method section, were confirmed by these experimental results.}
  \textcolor{black}{This convergence behavior refers not only to the PSO inner loop but also to the closed-loop joint optimization paradigm in Section 3.2, which converges within two iterations as confirmed by the per-iteration results in Table \ref{table_gt_validation}.}
  Another important feature of GTLLF is its flexible design. For example, GTLLF can adopt different LLMs (ChatGPT or DeepSeek) for POI attribute matching (Table \ref{tab:compare_1}).
  Both options produced smaller errors than the other comparison methods, and within this group, the ChatGPT-based approach performed better than DeepSeek.

  \textcolor{black}{As shown in Table \ref{table_compare_2}, TPS exhibited pathological degradation in multi-city experiments.
  The mean residual of TPS reached large values because the global smoothness constraint caused noise and outliers in LLM-matched data to propagate across the entire interpolation surface.
  Consequently, a subset of points was corrected to distant locations \textcolor{black}{(see Table S2)}.
  In contrast, GTLLF achieved reduced residuals because it combined an adaptive quality-control mechanism with local PSO-based refinement to minimize overfitting and prevent error propagation.
  The complete per-city residual statistics are provided in the Supplementary Material.}

\begin{table}[!h]
  \centering
  \color{black}
  \caption{The comparison of residuals for different methods.
  \textcolor{black}{Statistics were aggregated from per-city results. Detailed results for all cities are provided in Table S2.}}
  \setlength{\tabcolsep}{8pt}
  \resizebox{0.75\columnwidth}{!}{
    \begin{tabular}{lccc}
    \toprule
    \textbf{Method} & \textbf{Mean (m)} & \textbf{Std (m)} & \textbf{Median (m)} \\
    \midrule
    Origin            & 1286.74 & 2.94  & 1286.68 \\
    TPS           & 17.65   & 9.87  & 15.58   \\
    ETT               & 6.35    & 3.92  & 5.49    \\
    \textbf{GTLLF (Ours)} & \textbf{4.58} & \textbf{2.82} & \textbf{3.91} \\
    \bottomrule
  \end{tabular}}\label{table_compare_2}
\end{table}

  \textcolor{black}{Table~\ref{table_compare_2} also reports the standard deviation of per-city residuals.
  The mean residual of GTLLF ($4.58 \pm 2.82$~m) was lower than that of ETT ($6.35 \pm 3.92$~m), TPS ($17.65 \pm 9.87$~m), and the origin offset ($1286.74 \pm 2.94$~m).
  GTLLF reduced the mean residual by 1.77~m compared with ETT, by 13.07~m compared with TPS, and by 1282.16~m compared with the origin offset.
  The relatively small standard deviation of GTLLF (2.82~m) indicates that the method maintained consistently low residuals across all tested cities, whereas TPS exhibited high variability (9.87~m) due to its pathological degradation in certain cities.}

\begin{figure}[htb]
    \centering
    \color{black}
    \includegraphics[width=\linewidth]{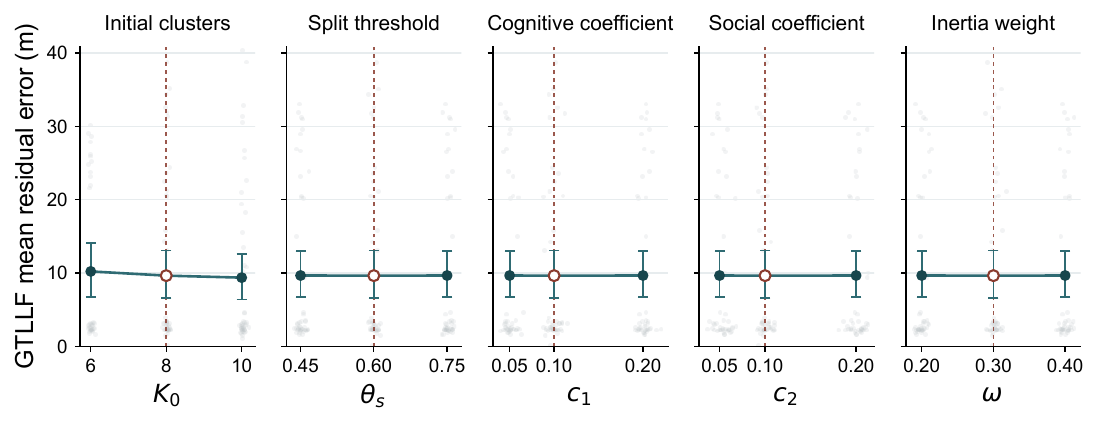}
\caption{Sensitivity analysis of hyperparameters on fusion residual Mean. (a) iterative self-organizing data analysis (ISODATA) $k_{\text{init}}$, (b) ISODATA split threshold, (c) particle swarm optimization (PSO) cognitive coefficient $c_{1}$, (d) PSO social coefficient $c_{2}$, (e) PSO inertia weight $\omega$.}
    \label{fig_sensitivity}
  \end{figure}

  \textcolor{black}{\subsection{Sensitivity analysis}
  To evaluate the robustness of GTLLF with respect to hyperparameter choices, we conducted a sensitivity analysis by varying key parameters across wide ranges: ISODATA initial cluster count $k_{\text{init}}$ from 6 to 10, split threshold from 0.45 to 0.75, PSO cognitive coefficient $c_{1}$ and social coefficient $c_{2}$ from 0.05 to 0.2, inertia weight $\omega$ from 0.2 to 0.4.
  As shown in Figure \ref{fig_sensitivity}, the fusion residual mean remained stable across all tested parameter ranges, with maximum variation within ±0.5 m (less than 5\% of the baseline residual), demonstrating that the framework is not sensitive to parameter tuning and operates robustly across diverse configurations.
  GTLLF maintained low residuals regardless of the specific hyperparameter settings, demonstrating that the framework was not sensitive to parameter tuning and operated robustly across diverse configurations.
  }

  {\color{black}{
  \subsection{Ground truth validation and ablation}
  To validate the reliability of the residual-based evaluation, a synthetic encrypted dataset was constructed by applying coordinate encryption to WGS-84 POI data from the Foursquare Open Source Places dataset\footnote{\url{https://opensource.foursquare.com/os-places/}}, where the original WGS-84 coordinates served as known ground truth.
  
  As discussed in the Introduction, direct accuracy validation on real data is infeasible due to the inaccessibility of decryption algorithms and the absence of public encrypted-to-WGS-84 POI pairs. 
  
  The WGS-84 coordinates, which represent physically measured positions without encryption induced offsets and thus serve as the known ground truth, are converted to BD-09 via the one-way WGS-84 to BD-09 empirical formula, and random Gaussian perturbations ($\sigma$ corresponding to 15 to 30 meters) are added in both $x$ and $y$ directions to simulate encryption offsets.
  For each shifted synthetic POI, all Baidu POIs within a 100-meter buffer are retrieved and the LLM identifies the best-matching one, whose name and address attributes are assigned to the synthetic POI, replicating the attribute ambiguity of real DSEP data.
  This ensures that the synthetic dataset simultaneously possesses both positional encryption and attribute ambiguity characteristics, allowing the fusion accuracy to be measured independently and directly without confounding factors.
  The synthetic encrypted POIs and their original WGS-84 counterparts were then processed through the proposed framework.
  ETT, TPS, and GTLLF were compared using the same metrics as Table \ref{tab:compare_1} on all cities where sufficient valid POI pairs were available.

  As shown in Table \ref{table_gt_validation}, GTLLF achieved the lowest average and median residuals among the three methods for both LLMs.
  For ChatGPT, GTLLF yielded an average residual of 37.36 meters and a median of 36.29 meters, outperforming ETT (37.55 and 36.47 meters) and TPS (293.98 and 297.13 meters).
  For DeepSeek-Chat, GTLLF achieved an average residual of 36.29 meters and a median of 35.20 meters, again outperforming both ETT and TPS.
  The residuals on the synthetic dataset (approximately 37 meters) were notably larger than those on real DSEP data (4.58 meters) because the synthetic data underwent encryption with additional random Gaussian perturbations superimposed, creating a substantially harder scenario than real single-encryption DSEP fusion.
  The narrow gap between GTLLF and ETT (0.19 meters) on this difficult synthetic dataset is expected: the added random noise component is irreducible by any deterministic transformation method, and both GTLLF and ETT converge toward the noise floor.
  \textcolor{black}{In the synthetic dataset with purely Gaussian noise, the global ETT already captures most of the systematic offset, leaving limited room for local PSO refinement.
  In real DSEP data, the encryption-induced nonlinear distortion is spatially heterogeneous and non-Gaussian, making local PSO refinement substantially more effective, as evidenced by the 1.77~m improvement on real data (Table \ref{table_compare_2}).}
  The key observation is that GTLLF still outperforms ETT while dramatically suppressing TPS degradation (293.98 meters for TPS), confirming its robustness even under extreme noise conditions.
  The comparison between ETT and GTLLF also constitutes an ablation study. ETT corresponds to the global step alone, using only the empirical trigonometric transformation with strict string matching, whereas GTLLF adds the proposed modules on top of ETT, namely LLM-driven attribute matching for more reliable correspondences and the PSO-based local refinement with quality control and closed-loop feedback. The residual difference between the two methods therefore isolates the contribution of these proposed components.
  The low residuals corresponded to high absolute accuracy, confirming that the residual metric used throughout this paper was a reliable indicator of fusion quality.
  \textcolor{black}{Table \ref{table_gt_validation} further reports the residuals of GTLLF after each iteration of the iterative loop. The residual decreased from Iter-1 to Iter-2 and remained stable thereafter, and the negligible difference between Iter-2 and Iter-3 indicates that the iterative loop essentially converged within two iterations. This behavior is consistent with the sample-purification and radius-contraction mechanism analyzed in Section \ref{sec_disc_loop}: once the platform-inconsistent pairs were filtered out in the first two iterations, the third iteration operated on an already-purified correspondence set and produced only marginal refinement.}

  A methodological caveat of this synthetic validation deserves explicit clarification.
  The ETT plays a dual role in this experiment, serving both as the generator of the synthetic encrypted coordinates and as the comparison baseline, which could raise an inverse crime (circular validation) concern since the baseline shares the transformation family used in data construction.
  Two factors mitigate this concern.
  The added random Gaussian perturbations (15 to 30 meters) destroy the exact invertibility of the encryption process, so even the generating transformation cannot recover the ground truth without residual error.
  GTLLF consistently outperformed ETT under identical noise conditions, indicating that the improvement originates from the closed-loop joint optimization paradigm rather than from fitting the specific transformation used in data generation.
  }}

  \begin{table}[!h]
  \centering
  \color{black}
  \caption{Ground truth validation of fusion accuracy using the synthetic encrypted dataset.
  For GTLLF, Iter-1, Iter-2 and Iter-3 correspond to the residuals after the first, second and third outer synergic iterations respectively, where Iter-3 is the reported final result.
  Bold values indicate the best performance among empirical trigonometric transformation (ETT), thin-plate spline (TPS) and global-to-local location fusion (GTLLF) for each LLM.}
  \label{table_gt_validation}
  \setlength{\tabcolsep}{6pt}
  \resizebox{0.85\linewidth}{!}{%
  \footnotesize
  \begin{tabular}{l l | c c}
  \toprule
  \textbf{LLM} & \textbf{Method}
   & \textbf{Average (m)} & \textbf{Median (m)} \\
  \midrule
  \multirow{6}{*}{ChatGPT}
   & Origin                                   & 1723.24 & 1723.01 \\
   & ETT                                      & 37.55   & 36.47   \\
   & TPS                                      & 293.98  & 297.13  \\
   & GTLLF (Iter-1)                           & 54.19   & 58.41   \\
   & GTLLF (Iter-2)                           & 37.36   & 36.29   \\
   & \textbf{GTLLF (Iter-3, Final)}           & \textbf{37.36} & \textbf{36.29} \\
  \midrule
  \multirow{6}{*}{DeepSeek-Chat}
   & Origin                                   & 1705.62 & 1708.02 \\
   & ETT                                      & 36.52   & 35.45   \\
   & TPS                                      & 92.68   & 78.73   \\
   & GTLLF (Iter-1)                           & 52.29   & 52.32   \\
   & GTLLF (Iter-2)                           & 36.29   & 35.20  \\
   & \textbf{GTLLF (Iter-3, Final)}           & \textbf{36.29} & \textbf{35.20} \\
  \bottomrule
  \end{tabular}
  }
  \end{table}

\subsection{The results of attribute fusion}
  After the GTLLF, with the accurate spatial location, location-synergic attribute fusion can be implemented.
  According to Table \ref{table2}, uncertainty level IV is the most uncertain, which is saved separately and judged to determine whether they are the corresponding DSEP. The other levels can be fused directly. 

\begin{figure}[htb]
    \centering
    \includegraphics[width=0.5\linewidth]{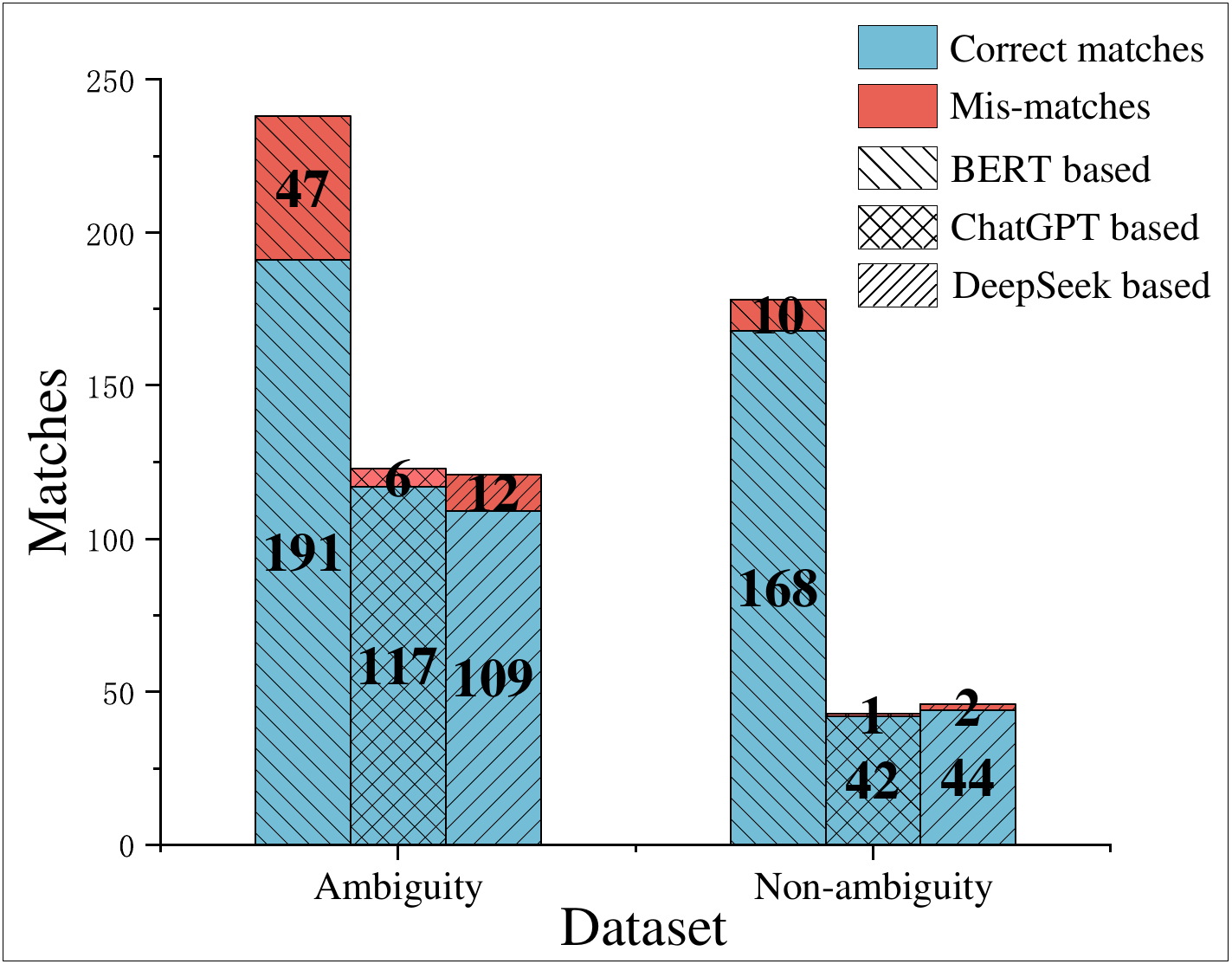}
    \caption{Attribute fusion evaluation results}
    \label{LLM_exp}
  \end{figure}

  In terms of matching accuracy, it can be indirectly evaluated through location error, as attribute fusion and location fusion are closely related.
  Attribute mismatches inevitably cause large residuals in coordinate transformation, meaning that accurate location fusion also reflects reliable POI matching, without the need for ground truth validation.
  The direct evaluation of attribute fusion is shown in Figure \ref{LLM_exp}.
  \textcolor{black}{The ambiguous and non-ambiguous subsets were divided according to the edit-distance similarity of the name and address strings, where pairs with similarity between 0.6 and 0.8 were regarded as ambiguous, and pairs with similarity above 0.9 or below 0.3 as non-ambiguous.}
  \textcolor{black}{The BERT baseline was fine-tuned on manually labeled POI pairs collected from five cities that were not included in the main test set, and the same name and address input format was used as for the LLM-based methods.}
  The ChatGPT-based method achieved the best performance, with an accuracy of 95.12\% on ambiguous data and 97.67\% on non-ambiguous data.
It performed significantly better than the BERT-based approach and slightly better than DeepSeek.
The BERT-based method produced a larger total number of correct matches compared with ChatGPT and DeepSeek.
  This is because the LLM-based methods require a higher confidence level to confirm a match, which improves matching reliability but results in fewer overall matches than BERT.

  \textcolor{black}{
  \subsection{LLM- and encrypted map-driven training-free LULC mapping}
  To validate the capability of the correction formula for remote sensing product georeferencing, we apply the proposed joint LLM- and encrypted POI-driven training-free LULC mapping (LEPF-LULC) to downstream LULC mapping across 31 provincial capitals and municipalities.}

\begin{figure}[htb]
    \centering
    \includegraphics[width=0.8\linewidth]{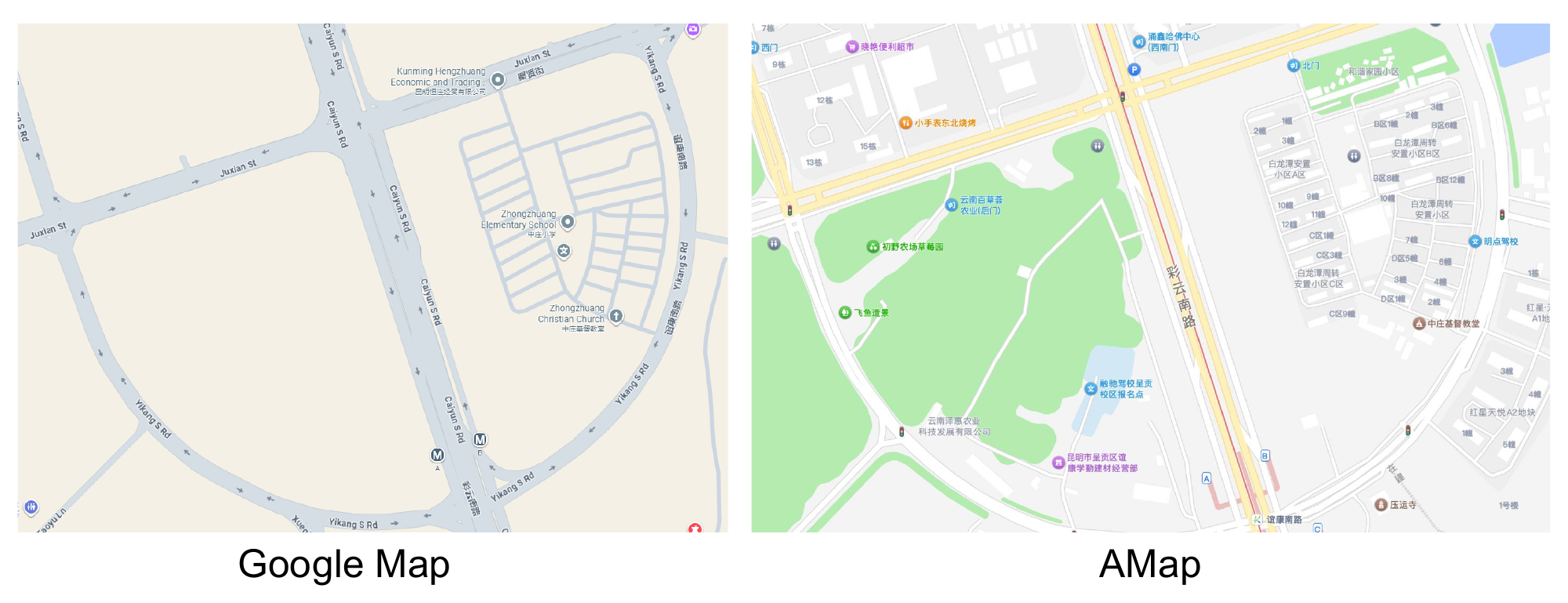}
\caption{Google Maps LULC data in Kunming, China. Within mainland China, Google Maps provides only basic categories and lacks fine-grained urban LULC elements for some cities, and its data updates are infrequent, limiting its applicability as a LULC reference.}
    \label{google_map}
  \end{figure}

  {\color{black}{
  Google Maps LULC data were not adopted as the reference because, within mainland China, they provide only basic categories such as road networks, water bodies, and terrain while lacking fine-grained urban LULC elements such as building parcels (Figure \ref{google_map}), and their API access restrictions and infrequent updates further limit data accessibility and reliability.
  OSM-LULC was therefore adopted as the reference value for its native WGS-84 coordinates, crowd-sourced updates, and independence from both LEPF-LULC and EULUC pipelines.
  \color{black}{EULUC-China 2.0 \cite{LI20253029} was selected as the comparative baseline because it is the only available dataset that comprehensively matches our study in terms of city coverage, temporal proximity, and data structure.}
  A reclassification guided by semantic consistency and spatial comparability yielded six comparable categories (built-up area, transportation, education, medical, sports and culture, green space); incompatible categories were excluded.
  All datasets were clipped, reprojected, and rasterized onto an identical grid to ensure metrics were computed on the same spatial units.
  }}

  As shown in Table \ref{table_lulc_quantitative}, \textcolor{black}{LEPF-LULC achieved an average mIoU of 75.14\% and an average Macro-F1 of 83.11\%, outperforming EULUC-China 2.0, which obtained an average mIoU of 58.58\% and a Macro-F1 of 72.01\%.
  EULUC-China 2.0 lacked coverage or produced near-zero metrics in 5 of the 31 cities; these cities were excluded from its mean calculation to avoid underestimating EULUC performance, resulting in $N$=26 for EULUC and $N$=31 for LEPF-LULC.
  The complete per-city results are provided in the Supplementary Material.}
  \textcolor{black}{The advantage of LEPF-LULC stems from its progressive processing pipeline, where color classification, category reorganization, spatial embedding, and evaluation benchmark construction collectively minimize error propagation.}
  \textcolor{black}{The closed-loop joint optimization paradigm ensures that purified attribute matching and hierarchical constraints improve the effective utilization of electronic map source data, enabling fine-grained urban land features to be more stably transformed into LULC representations.}
  
  \textcolor{black}{This training-free paradigm, requiring no field-surveyed GCPs but only sparse POIs and electronic map vectors, directly inherits classification from electronic maps via the correction formula, producing vector-raster integrated output without boundary discretization errors.}
  \textcolor{black}{EULUC-China 2.0 exhibited relatively lower accuracy in several cities, which may be associated with inconsistent temporal baselines, missing coverage in certain urban areas, and update lag in rapidly urbanizing regions.}
  \textcolor{black}{These factors collectively limit its timeliness and fine-detail expressiveness in localized urban spaces, where LEPF-LULC benefits from synchronized updates with open electronic maps.}
  \textcolor{black}{Because OSM-LULC is crowd-sourced, it inherently contains label noise and incomplete coverage that propagate into the evaluation metrics. This reference error affects both LEPF-LULC and EULUC-China 2.0 equally because both are evaluated against the same OSM-LULC baseline, so the relative comparison remains valid. The absolute mIoU values may be moderately underestimated, and the consistent margin by which LEPF-LULC outperforms EULUC-China 2.0 across most cities confirms that the observed advantage is not attributable to reference data artifacts.}

  \begin{table}[!h]
      \centering
      \color{black}
      \caption{Quantitative comparison of \textcolor{black}{land-use/land-cover (LULC)} classification accuracy between EULUC and the proposed \textcolor{black}{joint LLM- and encrypted POI-driven training-free LULC mapping (LEPF-LULC)}.
  Metrics are mean Intersection-over-Union (mIoU) and Macro-F1, reported as mean $\pm$ standard deviation (\%) across cities.
      \textcolor{black}{$N$ denotes the number of cities evaluated. EULUC results were computed over 26 cities where EULUC provides coverage; cities with no EULUC data were excluded from the mean calculation for fairness. Detailed results for all cities are provided in Table S3.}}
      \label{table_lulc_quantitative}
      \renewcommand{\arraystretch}{1.2}
      \resizebox{0.6\linewidth}{!}{%
      \begin{tabular}{lcc}
      \toprule
      \textbf{Method ($N$)} & \textbf{mIoU (\%)} & \textbf{Macro-F1 (\%)} \\
      \midrule
      \textbf{LEPF-LULC (Ours, $N$=31)} & \textbf{75.14 $\pm$ 5.74} & \textbf{83.11 $\pm$ 4.90}  \\
      EULUC ($N$=26) & 58.58 $\pm$ 6.83 & 72.01 $\pm$ 5.74  \\
      \bottomrule
      \end{tabular}
      }
  \end{table}

  \section{Discussion}
  \subsection{\textcolor{black}{Mechanism of closed-loop synergic optimization}}
  \label{sec_disc_loop}
  {\color{black}
  The accuracy gain of the proposed framework stems from a global-to-local strategy combined with an iterative closed-loop that progressively purifies the correspondence set and contracts the local search radius.
  The global registration provides an initial transformation that captures the dominant nonlinear distortion of the encrypted coordinate systems, while the local PSO refinement addresses spatially heterogeneous residuals that the global model cannot resolve.
  The LLM-driven attribute matching within adaptively partitioned subregions supplies higher-quality control points than strict string matching, because the LLM resolves semantic ambiguities that cause mismatches in traditional methods.
  The iterative loop further improves accuracy through a monotone sample-purification and radius-contraction process.
  In the first iteration, the local search radius derived from global residual statistics is relatively loose, so certain POI pairs that share highly similar names and addresses but originate from different platform-specific acquisition conventions are accepted by the LLM.
  This situation arises frequently in dual-source encrypted maps because Baidu and Amap adopt distinct acquisition strategies for the same real-world entity, for example placing the marker at the main entrance instead of the building centroid.
  Their attributes remain nearly identical because both platforms inherit the same official name and address, whereas their coordinates carry an entity-level offset that is not attributable to coordinate-system encryption.
  Such pairs are geometrically inconsistent yet semantically indistinguishable, so they cannot be identified in the first iteration.

  After the first attribute-synergic location fusion updates $\theta$, the geometric residuals of these platform-inconsistent pairs exceed the local three-sigma threshold, so location-synergic attribute fusion attenuates their attribute confidence through the fuzzy-mathematics credibility grading described in Section \ref{sec_35_fuzzy} and removes them from $\Pi$ once their credibility falls below the preset threshold.
  The residual dispersion $\hat{\sigma}$ therefore decreases, and the search radius in the second iteration contracts accordingly.
  Under the tightened radius, the residual pairs fall outside each other's search neighborhood or receive very low geometric weights, and the second-round LLM matching filters them out.

  From the third iteration onward, the objective sequence $\{ \mathcal{L}^{(k)} \}$ enters a saturation regime because all pairs whose credibility falls below the threshold have already been removed.
  The remaining set is stable, $\mathcal{P}^{(k)} = \mathcal{P}^{(k-1)}$ up to negligible boundary cases, so $\hat{\sigma}$ ceases to contract and the search radius stabilizes.
  The PSO update of $\theta$ operates on the same sample support and search bounds as in the previous iteration and yields only marginal parameter adjustments.
  The termination criterion in Formula \ref{eq_stop} is therefore triggered within two to three iterations in practice, validating the iterative loop as a bounded refinement mechanism rather than an open-ended one.

  In summary, the residual variance can be decomposed into a systematic component removed by the global transformation, a spatially heterogeneous component absorbed by the local PSO refinement, and an irreducible noise floor dominated by entity-level platform conventions.
  The closed-loop successively eliminates the first two components and contracts the feasible region accordingly, so the convergence within two iterations is the expected outcome of this variance decomposition rather than an empirical coincidence.
  }

  \subsection{\textcolor{black}{Efficiency, automation, and LLM reliability}}
  {\color{black}
  The local search circle strategy reduces the time complexity of POI matching from $O(N^2)$ to between $O(N)$ and $O(1.5N)$, cutting pairwise comparisons by over 93\% (Table \ref{tab:timeTable}).
  This reduction makes the framework applicable to million-scale POI scenarios, as the number of LLM queries and token consumption scale linearly with data volume.
  In terms of API cost, processing the filtered POI pairs using the ChatGPT API cost approximately \$50, while DeepSeek-Chat incurred even lower costs.
  Batch processing was performed with 10 POI pairs per query to ensure each request fit within the context window.
  }

  \begin{table}[h!]
    \centering
\caption{Comparison of \textcolor{black}{points of interest (POI)} pairwise computations.
    The proposed local search circle significantly reduces computations, demonstrating much higher efficiency compared to exhaustive direct matching and fixed distance matching with a predefined  search radius.}
    \resizebox{0.8\linewidth}{!}{%
    \begin{tabular}{cccc}
      \hline
      City       & Direct Matching \textcolor{black}{(times)} & Fixed Distance Matching \textcolor{black}{(times)} & Local Search Circle \textcolor{black}{(times)}      \\
      \hline
      Shanghai  & 46958   & 24.83     & \textbf{1.35 (94.56\%$\downarrow$)} \\
      Shenzhen  & 34069   & 32.58      & \textbf{1.34 (95.89\%$\downarrow$)} \\
      Guangzhou & 34758 &  24.00   & \textbf{1.45 (93.96\%$\downarrow$)} \\
      Average &36202 &27.14 & \textbf{1.38 (94.92\%$\downarrow$)}\\
      \bottomrule
    \end{tabular}
    }

    \label{tab:timeTable}
  \end{table}

  {\color{black}
  The framework operates with a high level of automation.
  The LAM leverages pre-trained LLMs whose parameters remain frozen throughout the pipeline, with no fine-tuning, no gradient updates, and no task-specific training data involved.
  Expert knowledge is integrated through prompt engineering rather than supervised learning, making the approach applicable in sample-poor scenarios where fine-tuned models such as BERT exhibit limited generalizability.
  The attribute-synergic location fusion was designed as a training-free pipeline where alignment was directly guided by attribute matching results without any manual intervention.
  The framework is model-agnostic and can incorporate any attribute similarity model, including LLMs, BERT, or traditional string similarity methods, as the initializer for attribute matching.

  Regarding LLM reliability, the attribute matching task is formulated as string similarity comparison rather than content generation, which inherently limits the hallucination risk.
  The confidence-based response design and fuzzy mathematics fusion mechanism further filter occasional erroneous outputs, ensuring reliable attribute matching across the dataset.
  Most importantly, the closed-loop paradigm provides an additional geometric back-check: if the LLM incorrectly matches two geographically distant POIs, the resulting large residual will automatically downgrade attribute confidence in the subsequent fuzzy filtering step, and the pair will be eliminated in the next iteration.
  This geometric validation mechanism makes the pipeline robust to occasional LLM errors.
  Experimental results show that the LLM-based approach achieved matching accuracies of 94\% to 99\%, outperforming a fine-tuned BERT model in ambiguous scenarios.
  }

  \subsection{\textcolor{black}{Validity of evaluation and downstream transfer}}
  {\color{black}
  The residual-based evaluation follows the same principle as bundle adjustment in photogrammetry, where reconstruction quality is evaluated by reprojection residuals without requiring independent ground truth for every point.
  A low residual indicates that two independently encrypted coordinate representations of the same geographic entity are geometrically consistent after transformation, which is the necessary and sufficient condition for reliable coordinate unification.
  This is distinct from absolute georeferencing accuracy, which was addressed separately by the synthetic ground-truth validation (Section 4.6).
  The residual metric is a reliable proxy for fusion quality because of a monotonic relationship between residual and matching precision.
  Incorrect matches produce large residuals following the spatial distribution of unrelated POIs, while correct matches produce small residuals reflecting only irreducible encryption noise.
  This relationship was empirically confirmed by the synthetic ground-truth validation (Table \ref{table_gt_validation}), where methods with lower residuals consistently achieved higher absolute accuracy.
  The residual metric was computed over all matched POI pairs rather than a sparse set of ground control points, providing a spatially comprehensive assessment that captures local distortions invisible to sparse GCP-based evaluation.

  The joint LLM- and encrypted POI-driven training-free LULC mapping demonstrates the downstream transferability of the proposed framework.
  The correction formula derived from DSEP fusion transforms encrypted vector data to WGS-84 coordinates, enabling direct overlay with remote sensing imagery.
  Each meter of positional residual corresponds to one to two pixels of georeferencing error in 0.5 to 1 m resolution imagery, so the 1.77 m improvement over the open-source baseline directly translates into alignment gains of approximately two to four pixels for mainstream high-resolution remote sensing products.
  Electronic maps employ consistent semantic-layer encoding, where a single POI within any semantic region suffices to determine the classification of all polygons sharing that layer.
  This characteristic makes the method particularly valuable in regions where field-surveyed GCPs are difficult to obtain.
  The approach requires no training, no GPU inference, and no annotated data, producing vector-raster integrated LULC mapping results with $O(N)$ time complexity.
  }

  \subsection{\textcolor{black}{Limitations and future work}}
  {\color{black}
  Several limitations should be noted.
  First, currently available public LULC datasets generally provide only coarse first-level category labels, lacking open-source datasets at second-level and third-level fine scales.
  Consequently, the quantitative comparison with EULUC-China 2.0 was conducted only at the coarse first-level category scale.
  When comparing multi-source data, imagery temporal phases should be kept as close as possible, as significant temporal differences may introduce additional errors.
  Constructing fine-grained reference datasets by integrating official planning documents and volunteered geographic information would enable validation at finer scales in future work.

  Second, the LULC dataset constructed in this study originates from encrypted map data, and its temporal currency may not match that of field survey data.
  Individual ground truth changes may not be reflected in a timely manner.
  The overall accuracy is sufficient for deep learning or weakly-supervised change detection tasks as training labels, as weakly-supervised learning frameworks inherently possess noise-suppression capabilities during training.
  Incorporating periodic change detection to refresh the inherited land use labels would further improve the temporal currency of the dataset.

  Third, the improvement of GTLLF over the open-source baseline ETT is incremental yet meaningful in absolute terms (1.77~m), as ETT already reduces the origin offset by approximately 99.5\%.
  All tested methods that incorporate local optimization converge to residuals within a narrow range, indicating that this behavior reflects the asymptotic performance ceiling of deterministic transformation methods under encrypted coordinate systems rather than a deficiency of the proposed method.
  Hybrid frameworks that combine deterministic transformation with stochastic residual denoising may push below this ceiling in future work.

  Fourth, the residual-based evaluation relies on the assumption that lower residuals correspond to higher fusion quality, which was validated on synthetic data but cannot be directly verified on real encrypted data due to the inaccessibility of decryption algorithms.
  In the synthetic validation, the ETT also serves as both the data generator and the comparison baseline, and although the added random perturbations and the consistent advantage of GTLLF mitigate this circularity, it cannot be fully excluded.
  Collecting partially labeled correspondences through field surveys would enable direct verification on real encrypted data.

  Fifth, the framework currently addresses fusion between two encrypted coordinate systems (BD-09 and GCJ-02).
  Extension to three or more encrypted sources would require additional pairwise fusion steps, and the scalability of the closed-loop joint optimization paradigm to such settings remains to be investigated.
  Generalizing the closed-loop formulation to multi-source encrypted fusion is therefore left as future work.
  }

  \section{Conclusion}
  {\color{black}
  To the best of our knowledge, this study is the first to propose a training-free, reference-free, and highly automated closed-loop joint optimization paradigm for DSEP fusion.
  The paradigm, termed the closed-loop joint optimization paradigm, alternates attribute matching and location optimization in a bidirectional feedback loop that progressively purifies the correspondence set and essentially converges within two iterations.
  Within this paradigm, the LLM-driven attribute matching establishes correspondences through multi-round dialogues without any training, the global-to-local location fusion optimizes the transformation coefficients through the improved PSO algorithm within ISODATA-clustered subregions, and the LLM-fuzzy method reassesses attribute confidence conditioned on the updated geometric residuals.
  Experiments across 31 provincial capitals and municipalities in mainland China demonstrated that the proposed method achieved an average location fusion residual of 4.58 meters and an attribute fusion accuracy of 95.12\%, outperforming the open-source baseline and SOTA with improvements of 1.77 meters and 14.87\%, respectively.
  The joint LLM- and encrypted POI-driven training-free LULC mapping method further achieved an average mIoU of 75.14\% and an average Macro-F1 of 83.11\%, demonstrating effective alignment of encrypted vector data with WGS-84 reference data.

  The closed-loop joint optimization paradigm transfers beyond POI fusion to remote sensing applications, where the same interdependent fusion principle enables training-free vector-raster integrated LULC mapping by aligning encrypted electronic-map vectors with WGS-84 imagery.
  The proposed reference-free evaluation method supports stage-wise quality monitoring for large-scale automated fusion pipelines.
  The high-accuracy georeferenced LULC dataset covers major urban built-up areas nationwide, providing geometrically accurate base data for remote sensing interpretation and GIS mapping.
  Future work will focus on multi-source encrypted fusion, finer-grained LULC classification, and direct verification on real encrypted data.
  }

\section*{CRediT authorship contribution statement}
\textbf{Chang Li}: Writing -- review and editing, Conceptualization, Methodology, Supervision, Project administration, Funding acquisition, Resources.
\textbf{Xingtao Peng}:  Data curation, Formal analysis, Investigation, Software, Validation, Visualization, Writing -- original draft.
\textbf{Yongjun Zhang}: Supervision, Resources.
\textbf{Yinfei He}: Data curation, Software, Validation.
\textbf{Cairun Huang}: Investigation, Visualization.

\section*{Acknowledgements}
The authors are grateful for the comments and contributions of the editors, anonymous reviewers and the members of the editorial team. This work was supported by the Key Program of the National Natural Science Foundation of China under Grant Nos. 42030102, the National Natural Science Foundation of China (NSFC) under Grant Nos. 41771493 and 41101407, and the Fundamental Research Funds for the Central Universities under Grant CCNU25JCPT001 and CCNU22QN019. 

\section*{Data availability statement}
The dataset that supports the findings of this study is available with the identifier at the link: https://figshare.com/s/7e20667e7cd9bb85cea0
  \bibliographystyle{elsarticle-num-names}
  \bibliography{bibliography}

@incollection{almeidaAutomaticPOIMatching2018a,
  title     = {Automatic POI Matching Using an Outlier Detection Based Approach},
  booktitle = {Advances in Intelligent Data Analysis XVII},
  author    = {Almeida, Alexandre and Alves, Ana and Gomes, Rui},
  editor    = {Duivesteijn, Wouter and Siebes, Arno and Ukkonen, Antti},
  year      = 2018,
  volume    = {11191},
  pages     = {40--51},
  publisher = {Springer International Publishing},
  address   = {Cham},
  doi       = {10.1007/978-3-030-01768-2_4},
  urldate   = {2024-08-26},
  isbn      = {978-3-030-01767-5 978-3-030-01768-2}
}

@article{ballISODATANovelMethod1965a,
  title   = {ISODATA, a Novel Method of Data Analysis and Pattern Classification},
  author  = {Ball, Geoffrey H.},
  year    = 1965,
  journal = {stanford research institute},
  pages   = {AD--699616},
  urldate = {2024-08-23}
}

@misc{brownLanguageModelsAre2020,
  title         = {Language Models Are Few-Shot Learners},
  author        = {Brown, Tom B. and Mann, Benjamin and Ryder, Nick and Subbiah, Melanie and Kaplan, Jared and Dhariwal, Prafulla and Neelakantan, Arvind and Shyam, Pranav and Sastry, Girish and Askell, Amanda and Agarwal, Sandhini and {Herbert-Voss}, Ariel and Krueger, Gretchen and Henighan, Tom and Child, Rewon and Ramesh, Aditya and Ziegler, Daniel M. and Wu, Jeffrey and Winter, Clemens and Hesse, Christopher and Chen, Mark and Sigler, Eric and Litwin, Mateusz and Gray, Scott and Chess, Benjamin and Clark, Jack and Berner, Christopher and McCandlish, Sam and Radford, Alec and Sutskever, Ilya and Amodei, Dario},
  year          = 2020,
  month         = jul,
  number        = {arXiv:2005.14165},
  eprint        = {2005.14165},
  primaryclass  = {cs},
  publisher     = {arXiv},
  doi           = {10.48550/arXiv.2005.14165},
  urldate       = {2024-08-18},
  archiveprefix = {arXiv}
}

@article{caiResearchMultisourcePOI2022a,
  title   = {Research on Multi-Source POI Data Fusion Based on Ontology and Clustering Algorithms},
  author  = {Cai, Li and Zhu, Longhao and Jiang, Fang and Zhang, Yihan and He, Jing},
  year    = 2022,
  month   = mar,
  journal = {Applied Intelligence},
  volume  = {52},
  number  = {5},
  pages   = {4758--4774},
  issn    = {1573-7497},
  doi     = {10.1007/s10489-021-02561-6},
  urldate = {2024-08-29},
  langid  = {english}
}

@article{cousseauLinkingPlaceRecords2021a,
  title   = {Linking Place Records Using Multi-View Encoders},
  author  = {Cousseau, Vincius and Barbosa, Luciano},
  year    = 2021,
  month   = sep,
  journal = {Neural Computing and Applications},
  volume  = {33},
  number  = {18},
  pages   = {12103--12119},
  issn    = {0941-0643, 1433-3058},
  doi     = {10.1007/s00521-021-05932-9},
  urldate = {2024-08-29},
  langid  = {english}
}

@misc{deepseek-aiDeepSeekV3TechnicalReport2025,
  title         = {DeepSeek-V3 Technical Report},
  author        = {{DeepSeek-AI} and Liu, Aixin and Feng, Bei and Xue, Bing and Wang, Bingxuan and Wu, Bochao and Lu, Chengda and Zhao, Chenggang and Deng, Chengqi and Zhang, Chenyu and Ruan, Chong and Dai, Damai and Guo, Daya and Yang, Dejian and Chen, Deli and Ji, Dongjie and Li, Erhang and Lin, Fangyun and Dai, Fucong and Luo, Fuli and Hao, Guangbo and Chen, Guanting and Li, Guowei and Zhang, H. and Bao, Han and Xu, Hanwei and Wang, Haocheng and Zhang, Haowei and Ding, Honghui and Xin, Huajian and Gao, Huazuo and Li, Hui and Qu, Hui and Cai, J. L. and Liang, Jian and Guo, Jianzhong and Ni, Jiaqi and Li, Jiashi and Wang, Jiawei and Chen, Jin and Chen, Jingchang and Yuan, Jingyang and Qiu, Junjie and Li, Junlong and Song, Junxiao and Dong, Kai and Hu, Kai and Gao, Kaige and Guan, Kang and Huang, Kexin and Yu, Kuai and Wang, Lean and Zhang, Lecong and Xu, Lei and Xia, Leyi and Zhao, Liang and Wang, Litong and Zhang, Liyue and Li, Meng and Wang, Miaojun and Zhang, Mingchuan and Zhang, Minghua and Tang, Minghui and Li, Mingming and Tian, Ning and Huang, Panpan and Wang, Peiyi and Zhang, Peng and Wang, Qiancheng and Zhu, Qihao and Chen, Qinyu and Du, Qiushi and Chen, R. J. and Jin, R. L. and Ge, Ruiqi and Zhang, Ruisong and Pan, Ruizhe and Wang, Runji and Xu, Runxin and Zhang, Ruoyu and Chen, Ruyi and Li, S. S. and Lu, Shanghao and Zhou, Shangyan and Chen, Shanhuang and Wu, Shaoqing and Ye, Shengfeng and Ye, Shengfeng and Ma, Shirong and Wang, Shiyu and Zhou, Shuang and Yu, Shuiping and Zhou, Shunfeng and Pan, Shuting and Wang, T. and Yun, Tao and Pei, Tian and Sun, Tianyu and Xiao, W. L. and Zeng, Wangding and Zhao, Wanjia and An, Wei and Liu, Wen and Liang, Wenfeng and Gao, Wenjun and Yu, Wenqin and Zhang, Wentao and Li, X. Q. and Jin, Xiangyue and Wang, Xianzu and Bi, Xiao and Liu, Xiaodong and Wang, Xiaohan and Shen, Xiaojin and Chen, Xiaokang and Zhang, Xiaokang and Chen, Xiaosha and Nie, Xiaotao and Sun, Xiaowen and Wang, Xiaoxiang and Cheng, Xin and Liu, Xin and Xie, Xin and Liu, Xingchao and Yu, Xingkai and Song, Xinnan and Shan, Xinxia and Zhou, Xinyi and Yang, Xinyu and Li, Xinyuan and Su, Xuecheng and Lin, Xuheng and Li, Y. K. and Wang, Y. Q. and Wei, Y. X. and Zhu, Y. X. and Zhang, Yang and Xu, Yanhong and Xu, Yanhong and Huang, Yanping and Li, Yao and Zhao, Yao and Sun, Yaofeng and Li, Yaohui and Wang, Yaohui and Yu, Yi and Zheng, Yi and Zhang, Yichao and Shi, Yifan and Xiong, Yiliang and He, Ying and Tang, Ying and Piao, Yishi and Wang, Yisong and Tan, Yixuan and Ma, Yiyang and Liu, Yiyuan and Guo, Yongqiang and Wu, Yu and Ou, Yuan and Zhu, Yuchen and Wang, Yuduan and Gong, Yue and Zou, Yuheng and He, Yujia and Zha, Yukun and Xiong, Yunfan and Ma, Yunxian and Yan, Yuting and Luo, Yuxiang and You, Yuxiang and Liu, Yuxuan and Zhou, Yuyang and Wu, Z. F. and Ren, Z. Z. and Ren, Zehui and Sha, Zhangli and Fu, Zhe and Xu, Zhean and Huang, Zhen and Zhang, Zhen and Xie, Zhenda and Zhang, Zhengyan and Hao, Zhewen and Gou, Zhibin and Ma, Zhicheng and Yan, Zhigang and Shao, Zhihong and Xu, Zhipeng and Wu, Zhiyu and Zhang, Zhongyu and Li, Zhuoshu and Gu, Zihui and Zhu, Zijia and Liu, Zijun and Li, Zilin and Xie, Ziwei and Song, Ziyang and Gao, Ziyi and Pan, Zizheng},
  year          = 2025,
  month         = feb,
  number        = {arXiv:2412.19437},
  eprint        = {2412.19437},
  primaryclass  = {cs},
  publisher     = {arXiv},
  doi           = {10.48550/arXiv.2412.19437},
  urldate       = {2025-04-06},
  archiveprefix = {arXiv}
}

@article{liDeepLearningMethod2022,
  title     = {Deep Learning Method for Chinese Multisource Point of Interest Matching},
  author    = {Li, Pengpeng and Liu, Jiping and Luo, An and Wang, Yong and Zhu, Jun and Xu, Shenghua},
  year      = 2022,
  journal   = {Computers, Environment and Urban Systems},
  volume    = {96},
  pages     = {101821},
  publisher = {Elsevier}
}

@article{liDifferentSourcingPoint2020,
  title     = {Different Sourcing Point of Interest Matching Method Considering Multiple Constraints},
  author    = {Li, Chengming and Liu, Li and Dai, Zhaoxin and Liu, Xiaoli},
  year      = 2020,
  journal   = {ISPRS International Journal of Geo-Information},
  volume    = {9},
  number    = {4},
  pages     = {214},
  publisher = {MDPI},
  urldate   = {2024-09-07},
  langid    = {american}
}

@article{liEnhancedSemanticRepresentation2023,
  title   = {Enhanced Semantic Representation Model for Multisource Point of Interest Attribute Alignment},
  author  = {Li, Pengpeng and Wang, Yong and Liu, Jiping and Luo, An and Xu, Shenghua and Zhang, Zhiran},
  year    = 2023,
  month   = oct,
  journal = {Information Fusion},
  volume  = {98},
  pages   = {101852},
  issn    = {1566-2535},
  doi     = {10.1016/j.inffus.2023.101852},
  urldate = {2025-04-06}
}

@article{linDeepLearningArchitecture2020a,
  title   = {A Deep Learning Architecture for Semantic Address Matching},
  author  = {Lin, Yue and Kang, Mengjun and Wu, Yuyang and Du, Qingyun and Liu, Tao},
  year    = 2020,
  month   = mar,
  journal = {International Journal of Geographical Information Science},
  volume  = {34},
  number  = {3},
  pages   = {559--576},
  issn    = {1365-8816, 1362-3087},
  doi     = {10.1080/13658816.2019.1681431},
  urldate = {2024-08-26},
  langid  = {english}
}

@inproceedings{maMultisourcePointofinterestMatching2025,
  title     = {Multisource Point-of-Interest Matching Method Based on Multi-Attribute Feature Similarity},
  booktitle = {Third International Conference on Geographic Information and Remote Sensing Technology (GIRST 2024)},
  author    = {Ma, Shiji and Guo, Li and Ren, Fang and Cui, Qifan and Yu, Zheng},
  year      = 2025,
  month     = apr,
  volume    = {13551},
  pages     = {165--177},
  publisher = {SPIE},
  doi       = {10.1117/12.3059736},
  urldate   = {2025-04-30}
}

@article{novackGraphbasedMatchingPointsofinterest2018a,
  title     = {Graph-Based Matching of Points-of-Interest from Collaborative Geo-Datasets},
  author    = {Novack, Tessio and Peters, Robin and Zipf, Alexander},
  year      = 2018,
  journal   = {ISPRS International Journal of Geo-Information},
  volume    = {7},
  number    = {3},
  pages     = {117},
  publisher = {MDPI},
  urldate   = {2024-09-07}
}

@article{piechAutomaticPointsInterest2020,
  title     = {Towards Automatic Points of Interest Matching},
  author    = {Piech, Mateusz and {Smywinski-Pohl}, Aleksander and Marcjan, Robert and Siwik, Leszek},
  year      = 2020,
  journal   = {ISPRS International Journal of Geo-Information},
  volume    = {9},
  number    = {5},
  pages     = {291},
  publisher = {MDPI},
  urldate   = {2024-08-29}
}

@inproceedings{qiangMomentumContrastiveLearning2024,
  title     = {A Momentum Contrastive Learning Framework for Query-POI Matching},
  booktitle = {2024 IEEE International Conference on Data Mining (ICDM)},
  author    = {Qiang, Yuting and Zheng, Jianbing and Wu, Lixia and Wen, Haomin and Lou, Junhong and Deng, Minhui},
  year      = 2024,
  month     = dec,
  pages     = {833--838},
  issn      = {2374-8486},
  doi       = {10.1109/ICDM59182.2024.00101},
  urldate   = {2025-04-30},
  langid    = {american}
}

@article{qiuDeepNeuralNetwork2024,
  title     = {A Deep Neural Network Model for Chinese Toponym Matching with Geographic Pre-Training Model},
  author    = {Qiu, Qinjun and Zheng, Shiyu and Tian, Miao and Li, Jiali and Ma, Kai and Tao, Liufeng and Xie, Zhong},
  year      = 2024,
  month     = dec,
  journal   = {International Journal of Digital Earth},
  publisher = {Taylor \& Francis},
  issn      = {1753-8947},
  urldate   = {2024-10-04},
  copyright = {\copyright{} 2024 The Author(s). Published by Informa UK Limited, trading as Taylor \& Francis Group},
  langid    = {english}
}

@article{songAreAllCities2018a,
  title      = {Are All Cities with Similar Urban Form or Not? Redefining Cities with Ubiquitous Points of Interest and Evaluating Them with Indicators at City and Block Levels in China},
  shorttitle = {Are All Cities with Similar Urban Form or Not?},
  author     = {Song, Yongze and Long, Ying and Wu, Peng and Wang, Xiangyu},
  year       = 2018,
  month      = dec,
  journal    = {International Journal of Geographical Information Science},
  volume     = {32},
  number     = {12},
  pages      = {2447--2476},
  issn       = {1365-8816, 1362-3087},
  doi        = {10.1080/13658816.2018.1511793},
  urldate    = {2024-08-26},
  langid     = {english}
}

@article{sunConflatingPointInterest2023,
  title      = {Conflating Point of Interest (POI) Data: A Systematic Review of Matching Methods},
  shorttitle = {Conflating Point of Interest (POI) Data},
  author     = {Sun, Kai and Hu, Yingjie and Ma, Yue and Zhou, Ryan Zhenqi and Zhu, Yunqiang},
  year       = 2023,
  journal    = {Computers, Environment and Urban Systems},
  volume     = {103},
  pages      = {101977},
  publisher  = {Elsevier},
  doi        = {10.1016/j.compenvurbsys.2023.101977},
  urldate    = {2024-08-29}
}

@misc{sunMassiveActivationsLarge2024,
  title         = {Massive Activations in Large Language Models},
  author        = {Sun, Mingjie and Chen, Xinlei and Kolter, J. Zico and Liu, Zhuang},
  year          = 2024,
  month         = aug,
  number        = {arXiv:2402.17762},
  eprint        = {2402.17762},
  primaryclass  = {cs},
  publisher     = {arXiv},
  doi           = {10.48550/arXiv.2402.17762},
  urldate       = {2025-05-02},
  archiveprefix = {arXiv}
}

@article{toblerComputerMovieSimulating1970a,
  title      = {A Computer Movie Simulating Urban Growth in the Detroit Region},
  author     = {Tobler, W. R.},
  year       = 1970,
  month      = jun,
  journal    = {Economic Geography},
  volume     = {46},
  eprint     = {143141},
  eprinttype = {jstor},
  pages      = {234},
  issn       = {00130095},
  doi        = {10.2307/143141},
  urldate    = {2024-05-04}
}

@article{wangEfficientAlgorithmSpatiotextual2020,
  title   = {An Efficient Algorithm for Spatio-Textual Location Matching},
  author  = {Wang, Ning and Zeng, Jianping and Chen, Mingming and Zhu, Shunzhi},
  year    = 2020,
  month   = sep,
  journal = {Distributed and Parallel Databases},
  volume  = {38},
  number  = {3},
  pages   = {649--666},
  issn    = {1573-7578},
  doi     = {10.1007/s10619-020-07289-9},
  urldate = {2025-04-30},
  langid  = {english}
}

@article{wangGPTLargeLanguage2024,
  title      = {GPT, Large Language Models (LLMs) and Generative Artificial Intelligence (GAI) Models in Geospatial Science: A Systematic Review},
  shorttitle = {GPT, Large Language Models (LLMs) and Generative Artificial Intelligence (GAI) Models in Geospatial Science},
  author     = {Wang, Siqin and Hu, Tao and Xiao, Huang and Li, Yun and Zhang, Ce and Ning, Huan and Zhu, Rui and Li, Zhenlong and Ye, Xinyue},
  year       = 2024,
  month      = dec,
  journal    = {International Journal of Digital Earth},
  volume     = {17},
  number     = {1},
  pages      = {2353122},
  publisher  = {Taylor \& Francis},
  issn       = {1753-8947},
  doi        = {10.1080/17538947.2024.2353122},
  urldate    = {2024-10-04}
}

@misc{wuDeviationChinaMap,
  title        = {The Deviation of China Map as a Regression Problem},
  author       = {Wu, Yongzheng},
  urldate      = {2010-01-22},
  year         = 2010,
  howpublished = {https://wuyongzheng.github.io/china-map-deviation/paper.html}
}

@inproceedings{xingLocalPOIMatching2022,
  title     = {Local POI Matching Based on KNN and LightGBM Method},
  booktitle = {2022 2nd International Conference on Computer Science, Electronic Information Engineering and Intelligent Control Technology (CEI)},
  author    = {Xing, Xiaoqi and Lin, HaoJun and Zhao, FeiYun and Qiang, Shengzhou},
  year      = 2022,
  pages     = {455--458},
  publisher = {IEEE},
  urldate   = {2024-08-29}
}

@article{yaoSensingSpatialDistribution2017a,
  title   = {Sensing Spatial Distribution of Urban Land Use by Integrating Points-of-Interest and Google Word2Vec Model},
  author  = {Yao, Yao and Li, Xia and Liu, Xiaoping and Liu, Penghua and Liang, Zhaotang and Zhang, Jinbao and Mai, Ke},
  year    = 2017,
  month   = apr,
  journal = {International Journal of Geographical Information Science},
  volume  = {31},
  number  = {4},
  pages   = {825--848},
  issn    = {1365-8816, 1362-3087},
  doi     = {10.1080/13658816.2016.1244608},
  urldate = {2024-08-26},
  langid  = {english}
}

@inproceedings{yuSelfconsistentDeepGeometric2024,
  title     = {Self-Consistent Deep Geometric Learning for Heterogeneous Multi-Source Spatial Point Data Prediction},
  booktitle = {Proceedings of the 30th ACM SIGKDD Conference on Knowledge Discovery and Data Mining},
  author    = {Yu, Dazhou and Gong, Xiaoyun and Li, Yun and Qiu, Meikang and Zhao, Liang},
  year      = 2024,
  month     = aug,
  series    = {KDD '24},
  pages     = {4001--4011},
  publisher = {Association for Computing Machinery},
  address   = {New York, NY, USA},
  doi       = {10.1145/3637528.3671737},
  urldate   = {2025-06-14},
  isbn      = {979-8-4007-0490-1}
}

@article{zhangBBGeoGPTFrameworkLearning2024,
  title      = {BB-GeoGPT: A Framework for Learning a Large Language Model for Geographic Information Science},
  shorttitle = {BB-GeoGPT},
  author     = {Zhang, Yifan and Wang, Zhiyun and He, Zhengting and Li, Jingxuan and Mai, Gengchen and Lin, Jianfeng and Wei, Cheng and Yu, Wenhao},
  year       = 2024,
  journal    = {Information Processing \& Management},
  volume     = {61},
  number     = {5},
  pages      = {103808},
  publisher  = {Elsevier},
  urldate    = {2024-10-04}
}

@article{zhangGeoGPTAssistantUnderstanding2024,
  title      = {GeoGPT: An Assistant for Understanding and Processing Geospatial Tasks},
  shorttitle = {GeoGPT},
  author     = {Zhang, Yifan and Wei, Cheng and He, Zhengting and Yu, Wenhao},
  year       = 2024,
  journal    = {International Journal of Applied Earth Observation and Geoinformation},
  volume     = {131},
  pages      = {103976},
  publisher  = {Elsevier},
  urldate    = {2024-10-04}
}

@article{zhaoPoiPointEntity2022,
  title     = {Poi Point Entity Matching and Fusion Based on Multi Similarity Calculation},
  author    = {Zhao, J. and Niu, X. and Cui, Y. and Zhao, Y. and Guo, M. and Zhang, R.},
  year      = 2022,
  month     = oct,
  journal   = {ISPRS Annals of the Photogrammetry, Remote Sensing and Spatial Information Sciences},
  volume    = {X-3-W2-2022},
  pages     = {87--92},
  publisher = {Copernicus GmbH},
  issn      = {2194-9042},
  doi       = {10.5194/isprs-annals-X-3-W2-2022-87-2022},
  urldate   = {2025-04-30},
  langid    = {english}
}

@article{fischlerRandomSampleConsensus1981,
  title      = {Random Sample Consensus: A Paradigm for Model Fitting with Applications to Image Analysis and Automated Cartography},
  shorttitle = {Random Sample Consensus},
  author     = {Fischler, Martin A. and Bolles, Robert C.},
  year       = 1981,
  month      = jun,
  journal    = {Commun. ACM},
  volume     = {24},
  number     = {6},
  pages      = {381--395},
  issn       = {0001-0782},
  doi        = {10.1145/358669.358692},
  urldate    = {2025-07-01}
}

@inproceedings{kennedyParticleSwarmOptimization1995a,
  title     = {Particle Swarm Optimization},
  booktitle = {Proceedings of ICNN'95-International Conference on Neural Networks},
  author    = {Kennedy, James and Eberhart, Russell},
  year      = 1995,
  volume    = {4},
  pages     = {1942--1948},
  publisher = {ieee},
  urldate   = {2025-11-26}
}

@misc{zhangBridgingSemanticsGeometry2025,
  title = {Bridging semantics and geometry: A decoupled LVLM–SAM framework for reasoning segmentation in optical remote sensing},
  journal = {ISPRS Journal of Photogrammetry and Remote Sensing},
  volume = {237},
  pages = {217-235},
  year = {2026},
  issn = {0924-2716},
  doi = {https://doi.org/10.1016/j.isprsjprs.2026.04.036},
  url = {https://www.sciencedirect.com/science/article/pii/S0924271626002091},
  author = {Xu Zhang and Junyao Ge and Yang Zheng and Kaitai Guo and Jimin Liang},
}

@article{baoVisionMambaRemote2025,
  title = {Vision Mamba in remote sensing: a comprehensive survey of techniques, applications and outlook},
  author = {Bao, Muyi and Lyu, Shuchang and Xu, Zhaoyang and Zhou, Huiyu and others},
  journal = {arXiv preprint arXiv:2505.00630},
  year = {2025}
}

@article{LI20253029,
title = {Enhanced mapping of essential urban land use categories in China (EULUC-China 2.0): integrating multimodal deep learning with multisource geospatial data},
journal = {Science Bulletin},
volume = {70},
number = {18},
pages = {3029-3041},
year = {2025},
issn = {2095-9273},
doi = {https://doi.org/10.1016/j.scib.2025.07.006},
url = {https://www.sciencedirect.com/science/article/pii/S2095927325007200},
author = {Ziming Li and Bin Chen and Yufei Huang and Han Wang and Yadian Wang and Yiming Yuan and Xuecao Li and Jing M. Chen and Bing Xu and Peng Gong},
}

@ARTICLE{11509353,
  author={Qin, Haiming and Zhou, Weiqi and Zhu, Zhe and Weng, Qihao},
  journal={IEEE Geoscience and Remote Sensing Magazine}, 
  title={Remote Sensing for Understanding the Vertical Dimension of Urban Landscapes: A review on data, products, methods, and prospects}, 
  year={2026},
  volume={},
  number={},
  pages={2-34},
  doi={10.1109/MGRS.2026.3684995}}
\end{document}